\documentclass{article}

    \PassOptionsToPackage{super, compress}{natbib}

\usepackage[preprint]{neurips_2022}

\usepackage{framed}
\usepackage{multirow,tabularx}
\usepackage{placeins}
\usepackage[utf8]{inputenc} 
\usepackage[T1]{fontenc}    
\usepackage{hyperref}       
\usepackage{url}            
\usepackage{booktabs}       
\usepackage{amsfonts}       
\usepackage{nicefrac}       
\usepackage{microtype}      
\usepackage{xcolor}         
\usepackage{amsmath}
\usepackage{graphicx}
\usepackage{fullpage}
\usepackage{caption}
\usepackage{subcaption}

\usepackage{wrapfig}
\usepackage{lscape}
\usepackage{rotating}
\usepackage{epstopdf}

\makeatletter

\usepackage{multirow}
\usepackage{subcaption}
\usepackage{lineno}
\usepackage[normalem]{ulem}

\renewcommand{\sout}[1]{}
\renewcommand{\textcolor}[2]{{#2}}

\title{Improving Clinician Performance in Classification of EEG Patterns on the Ictal-Interictal-Injury Continuum using Interpretable Machine Learning}

\author{%
  Alina Jade Barnett Ph.D.\thanks{Equal contribution / co-first authors}\\
  Department of Computer Science\\
  Duke University\\
  \texttt{alina.barnett@duke.edu} \\
  \And
  Zhicheng Guo$^\ast$\\
  Department of Electrical and Computer Engineering \\
  Duke University \\
  \texttt{zhicheng.guo@duke.edu} \\
  \AND
  Jin Jing Ph.D.$^\ast$ \\
  Harvard University, Beth Israel Deaconess Medical Center \\
  \texttt{jjing@bidmc.harvard.edu} \\
  \And
  Wendong Ge Ph.D. \\
  Harvard University, Massachusetts General Hospital \\
  \texttt{wendong.ge@mgh.harvard.edu} \\
  \And
  Peter W. Kaplan \\
  Johns Hopkins Medicine \\
  \texttt{pkaplan@jhmi.edu} \\
  \And
  Wan Yee Kong \\
  Harvard University, Beth Israel Deaconess Medical Center \\
  \texttt{wkong@bidmc.harvard.edu} \\
  \And
  Ioannis Karakis \\
  Emory University \\
  \texttt{ioannis.karakis@emory.edu} \\
  \And
  Aline Herlopian \\
  Yale University \\
  \texttt{aline.herlopian@yale.edu} \\
  \And
  Lakshman Arcot Jayagopal \\
  University of Nebraska Medical Center \\
  \texttt{l.arcotjayagopal@unmc.edu} \\
  \And
  Olga Taraschenko \\
  University of Nebraska Medical Center \\
  \texttt{olha.taraschenko@unmc.edu} \\
  \And
  Olga Selioutski \\
  University of Mississippi Medical Center \\
  \texttt{oselioutski@umc.edu} \\
  \And
  Gamaleldin Osman \\
  Mayo Clinic \\
  \texttt{osman.gamaleldin@mayo.edu} \\
  \And
  Daniel Goldenholz \\
  Harvard University, Beth Israel Deaconess Medical Center \\
  \texttt{daniel.goldenholz@bidmc.harvard.edu} \\
  \And
  Cynthia Rudin Ph.D.\thanks{Co-senior authors} \\
  Department of Computer Science \\
  Duke University\\
  \texttt{cynthia@cs.duke.edu} \\
  \And
  M. Brandon Westover M.D., Ph.D.$^\dagger$ \\
  Harvard University, Beth Israel Deaconess Medical Center, MGH McCance Center for Brain Health \\
  \texttt{bwestove@bidmc.harvard.edu} \\
}

\begin{document}

\maketitle
\section*{Corresponding author contact information}
Zhicheng Guo\\
Duke University Pratt School of Engineering \\
Box 90291 \\
Durham, NC, 27708, USA  \\
zhicheng.guo@duke.edu, 347-720-8974\\

\section*{Structured Abstract}

Background: In intensive care units (ICUs), critically ill patients are monitored with electroencephalography (EEG) to prevent serious brain injury. EEG monitoring is constrained by clinician availability and EEG interpretation can be subjective and prone to inter-observer variability. Automated deep learning systems for EEG could reduce human bias and accelerate the diagnostic process. However, black box deep learning models are untrustworthy, difficult to troubleshoot, and lack accountability in real-world applications, leading to a lack of trust and adoption by clinicians.\\

Methods: We developed an interpretable deep-learning system that accurately classifies six types of potentially harmful EEG activity (Seizure, LPD, GPD, LRDA, GRDA, other) while providing faithful case-based explanations of its predictions. The model was trained on 50,697 50-second continuous EEG samples from 2,711 ICU patients collected between July 2006 and March 2020 at Massachusetts General Hospital. EEG samples were labeled as one of six EEG patterns by 124 domain experts and trained annotators. Eight medical professionals with relevant backgrounds completed a study in which they categorized 100 EEG samples into the six label categories, once with and once without AI assistance. The assistive power of this interpretable system was evaluated by comparing user diagnostic accuracy with and without AI assistance. Model discriminatory performance was evaluated with AUROC (area under the receiver operating characteristic curve) and AUPRC (area under the precision-recall curve). Model interpretability was also measured with task-specific neighborhood agreement statistics. Separately, we visualized the latent space of the neural network using dimension reduction techniques to examine whether the ictal-interictal-injury continuum hypothesis is supported by data.\\

Results: The performance of all users significantly improved when provided with AI assistance compared to without. Mean user diagnostic accuracy improved from 47\% to 71\% ($p<0.04$). The model achieves AUROCs of 0.87, 0.93, 0.96, 0.92, 0.93, 0.80 for classes Seizure, LPD, GPD, LRDA, GRDA, and Other, respectively. This performance is statistically significantly higher than that of the corresponding uninterpretable (black box) model with $p<0.0001$. Videos traversing the ictal-interictal-injury manifold from dimension reduction give insight into the layout of EEG patterns within the network's latent space, and illuminate relationships between EEG patterns that were previously hypothesized but had not been shown explicitly before. These results indicate that the ictal-interictal-injury continuum hypothesis is supported by data.\\

Conclusions: Users showed significant pattern classification accuracy improvement with the assistance of this interpretable deep learning model. The interpretable design facilitates effective human-AI collaboration; this system may improve diagnosis and patient care in clinical settings. It may also provide a better understanding of how EEG patterns relate to each other along the ictal-interictal-injury continuum.

\section*{Introduction \textcolor{red}{\sout{(587/3000 words) (total 2882/3000 for main text)}}}
\setcounter{footnote}{0}
Seizures or status epilepticus are found in 20\% of patients with severe medical and neurologic illness who undergo brain monitoring with electroencephalography (EEG) because of altered mental status \cite{pmid10668693, pmid10478706}, and every hour of seizures detected on EEG further increases the risk of permanent disability or death \cite{De_Marchis2016, pmid24595203}. Other than seizures, intermediate seizure-like patterns of brain activity consisting of periodic discharges or rhythmic activity are even more common, occurring in nearly 40\% of patients undergoing EEG monitoring \cite{pmid26943901}. Two recent studies found evidence that, like seizures, this type of activity also increases the risk of disability and death if it persists for a prolonged period \cite{pmid34231244, arXiv:2203.04920[stat.ME]}. The ``ictal-interictal-injury continuum'' (IIIC) hypothesis proposed by Chiappa et al. \cite{pmid8978624} posits that these ambiguous brainwave activities along with seizures lie along a spectrum. Although the IIIC provides a conceptual framework for understanding these ambiguous but potentially harmful EEG patterns, it remains challenging to categorize these EEG patterns in real-world clinical settings. Until recently, manual review of the EEG has been the only method for quantifying IIIC EEG activities and patterns\footnote{In this paper, we use the terms ``seizure and seizure-like events'' and ``IIIC EEG patterns'' interchangeably.}, which suffers from subjectivity due to the ambiguous nature of these patterns \cite{pmid29139014,pmid8978624}. 

Recently, progress in deep learning and the availability of large EEG datasets made it possible to develop automated algorithms to detect and classify seizures \cite{dlseizure1, dlseizure2, dlseizure3, dlseizure4} and other EEG patterns \cite{older_eeg_paper}, with one recent model achieving a level of accuracy comparable to physician experts \cite{pmid33131680}. 
However, there is a lack of interpretability\footnote{\textcolor{red}{In this paper, being ``interpretable''  and ``interpretability'' is defined as the model's ability to explain their predictions in a way that humans can understand. ``Uninterpretable'' and  ``black box'' mean the opposite of ``interpretable'', where the model cannot provide any explanation of its decision-making.}} in many of the previous models' decision-making processes, which renders them unsuitable for assisting human practitioners with medical decision-making.
Specifically, uninterpretable (or ``black box'') models are prone to silent failures during clinical operations due to either poor generalization or over-reliance on trivial medically irrelevant features \cite{eyedeployment, chestxray}. These failures lead to misdiagnoses and increased risks for patients. As a result, the FDA and the European Union (through the General Data Protection Regulation) have published new requirements and guidelines calling for interpretability and explainability in AI used for medical applications \cite{xai3, xai1, xai2}. While explainability techniques such as Grad-CAM, SHAP, etc. \cite{gradcameegseizure1, gradcameegseizure2, filtervizLRP, shapvalues, featuremappermutationentropy} try to explain model decisions \textit{post hoc}\footnote{\textcolor{red}{``After the fact,'' meaning that the model architecture, development, and training are completed before applying methods to explain the model. \cite{rudin2019stop}}}, these methods only approximate model reasoning and, as a result, different methods often give conflicting \textcolor{red}{\sout{reasoning} explanations even when used on the same model and sample}. This contrasts with inherently interpretable methods, which have perfect fidelity; that is, their explanation exactly matches predictor network's \textcolor{red}{\sout{reasoning} underlying calculations}. 

Our objective is to build an AI assistive tool for IIIC EEG pattern classification to reduce human subjectivity and improve user accuracy in practice with an interpretability-focused approach. We aim to better assist clinicians in classifying EEG patterns accurately and reliably, which is the crucial first step in the EEG reading process \cite{pmid25457454}. We also hope to gain insight into the relationship between EEG patterns, and develop evidence related to the ``ictal-interictal-injury continuum'' (IIIC) hypothesis.
We introduce a novel interpretable deep-learning algorithm to classify seizures and rhythmic and periodic EEG patterns. We propose an explanation method named ``This EEG Looks Like That EEG,'' abbreviated as TEEGLLTEEG.
Our proposed interpretable algorithm outperforms the current black box IIIC EEG pattern classification state-of-the-art in both classification performance and interpretability metrics. To the best of our knowledge, this is the first work in developing an inherently interpretable model for EEG signals. We demonstrate the clinical utility of our model in a retrospective analysis wherein all eight users significantly improve in pattern categorization when provided with AI assistance. 
Additionally, we map the network's latent space into two dimensions using a dimension-reduction algorithm, revealing that EEG patterns within the IIIC, despite being given distinct class names, do not exist in isolated islands. Rather, each class is connected to every other class via a sequence of transitional intermediate patterns, which we demonstrate in a series of videos. This lends support to the IIIC hypothesis.

\begin{figure}
  \centering
     \includegraphics[width=0.9\linewidth]{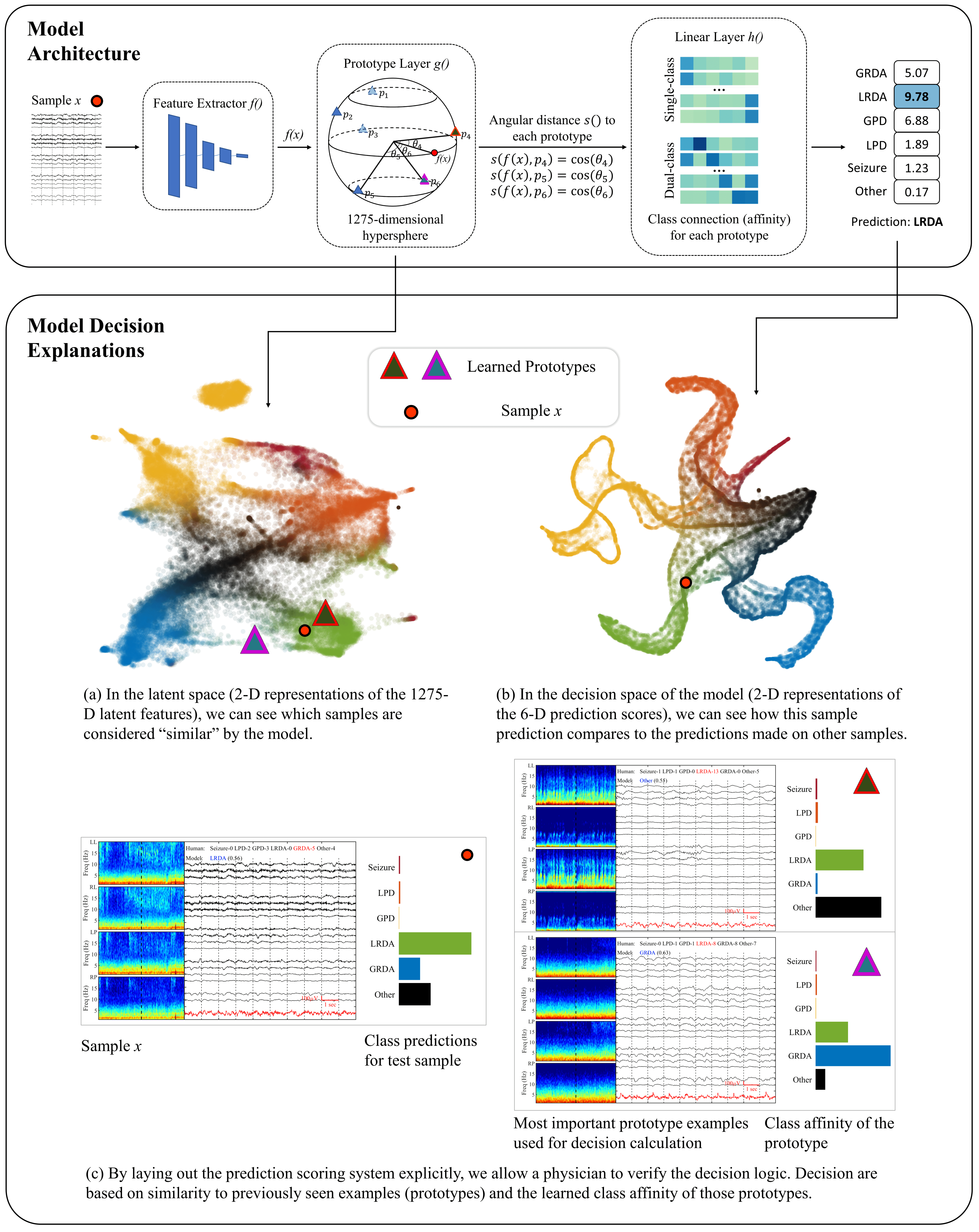}
     \caption{Upper: model architecture. Input sample $x$ is passed through a feature extractor $f()$, and a prototype layer $g()$ \textcolor{red}{(as in \citet{chen2019this})}. The prototype layer calculates angular distances \textcolor{red}{(as in \citet{donnelly2022deformable})} between the sample feature and the prototypes. The angular distances are multiplied with class affinity to generate the logits (class scores). The softmax calculation converts the logits into prediction probabilities. 
     Lower: three different ways an end user can see how the model reasons about the test sample: (a) Latent space explanations. (b) Decision space explanations. (c) Scoring system explanations.}
     \label{fig:3modes_of_interp}
\end{figure}

\begin{figure}
  \centering
     \includegraphics[width=\linewidth]{fig/Figure_GUI_recolored.png}
     \caption{Snapshot of the graphical user interface (GUI) of the interpretable system. The GUI integrates the three explanation modes detailed in Figure \ref{fig:3modes_of_interp}. This snapshot has minor simplifications for ease of reading. Full details and further information on the GUI layout can be found in Appendix \ref{app:gui}.}
     \label{fig:guisnapshot}
\end{figure}

\section*{Methods \textcolor{red}{\sout{(992/3000 words)}}}

\subsection*{EEG Data and Expert Labels}
Our network, called ProtoPMed-EEG, was trained and tested on a large-scale EEG study \cite{iir_eeg_paper} consisting of 50,697 events from 2,711 patients hospitalized between July 2006 and March 2020 who underwent continuous EEG as part of clinical care at Massachusetts General Hospital. EEG electrodes were placed according to the International 10-20 system. The large group was intended to ensure broad coverage of all variations of IIIC events encountered in practice. 124 EEG raters from 18 centers labeled the center 10-second of 50-second EEG segments. Raters produced one of the six labels: seizure (SZ), lateralized periodic discharges (LPD), generalized periodic discharges (GPD), lateralized rhythmic delta activity (LRDA), generalized rhythmic delta activity (GRDA), and ``Other'' patterns. ``Other'' included all patterns (including “normal”) other than seizures and the 4 rhythmic and periodic patterns (LPD, GPD, LRDA, GRDA). Patterns obscured by artifacts were scored by experts in the same way that this is done in clinical practice. That is when artifacts were present, but experts were still able to discern that one of the 5 target patterns was present, they were instructed to assign the target pattern as the label.
The data labeling procedure is described further in Appendix \ref{app:data}\textcolor{red}{\sout{ and Figure 10}}. Mean rater-to-rater inter-rater reliability (IRR) was moderate (percent agreement 52\%, $\kappa$ 42\%), and mean rater-to-majority IRR was substantial (percent agreement 65\%, $\kappa$ 61\%). As expert annotators do not always agree, we evaluated our model against the ``majority vote,'' such that the class selected by the most raters is the ground truth for each sample. The dataset was split into approximately equally sized training and test sets by patient ID to avoid leakage. Rather than allowing any training set sample to become a prototype, we limit our prototype candidates to 10641 samples that are thoroughly examined in the data labeling process ($\geq 20$ expert votes).

\subsection*{Interpretability through Model Design}
An overview of our novel model architecture design is shown in Figure \ref{fig:3modes_of_interp} (Upper). The model learns the feature extractor (initialized with weights from \citet{older_eeg_paper}), the prototype layer, and the final linear layer. In this work, the prototypes are divided into two categories, single-class prototypes and dual-class prototypes. Dual-class prototypes are novel to this work. Single-class prototypes represent EEG patterns that can be clearly attributed to one of the six classes we mentioned above. As described in the IIIC hypothesis, some EEG signals may exist in an intermediate state (e.g., in between LPD and seizure or in between GRDA and GPD); dual-class prototypes represent signals in such intermediate states. Each learned prototype corresponds to an actual EEG sample from the prototype subset. At test time, the latent feature of each input signal is compared against the learned prototypes by calculating their angular distances; the distances are passed through the last linear layer to produce prediction scores (logits). A more detailed version of the training process and the model architecture is provided in Appendix \ref{app:model_details} and Figure \ref{fig:detailed_arch}.

We show three modes of explanation provided by the model design in Figure \ref{fig:3modes_of_interp} (Lower): (a) Latent space explanations. (b) Decision space explanations. (c) Scoring system explanations. In this paper, we define interpretable models as models that ``explain their predictions in a way that humans can understand'' \cite{rudin2019stop}. Every prediction made by our model follows the same logic as the explanation provided by the model. This means that the model explanations have \textit{perfect fidelity} with the underlying decision-making process, by design. Figure \ref{fig:3modes_of_interp}(a) shows how the model perceives the test sample relative to previous cases by projecting the 1275-dimensional latent features to a human-comprehensible 2-dimensional space. Figure \ref{fig:3modes_of_interp}(b) shows the model's final classification of the test sample relative to the classifications of previous cases. Figure \ref{fig:3modes_of_interp}(c) shows how the model uses case-based reasoning to make its prediction (i.e., using previous examples to reason about a new case). This is achieved by learning a set of prototype samples that are representative of each single-class or dual-class category. Specifically, the model measures the similarity between a new case and the learned \textit{prototypical samples}. Each explanation is of the form ``this sample is class X because it is similar to these prototypes of class X, and not similar to prototypes of other classes.'' We call this method the TEEGLLTEEG explanation method because it makes explanations of the form ``this EEG looks like that EEG.'' 

The three modes of explanation are integrated into the final graphical user interface. We show a snap-shot of the dedicated GUI shown in Figure \ref{fig:guisnapshot}. Users could choose between explanation (a) and explanation (b) graphics in the drop-down menu; explanation (c) is presented as the top three prototypes on the right, with the scoring system visualized in color-coded pie charts. Here we only indicate components related to interpretability and explanation modes. For more details about the GUI, please see Appendix \ref{app:gui}.


\subsection*{User Study}
The potential clinical value of the proposed model was assessed with a multi-user study. The study cohort consists of a range of clinical practitioners including a nurse, an EEG technician, and MDs pursuing or having completed residency or fellowship training in neurology, stroke, dermatology, and neurophysiology. None of the participants have expertise in machine learning, and none were EEG experts (i.e., none were physicians who had completed clinical neurophysiology fellowship training). They represent an ideal cohort representing the realistic user population, as the interpretable system aims to assist clinical practitioners without any prerequisites for AI knowledge. Qualification and user backgrounds are further detailed in Appendix \ref{sec:user_info}. 

\begin{figure}
  \centering
  \includegraphics[width=\linewidth]{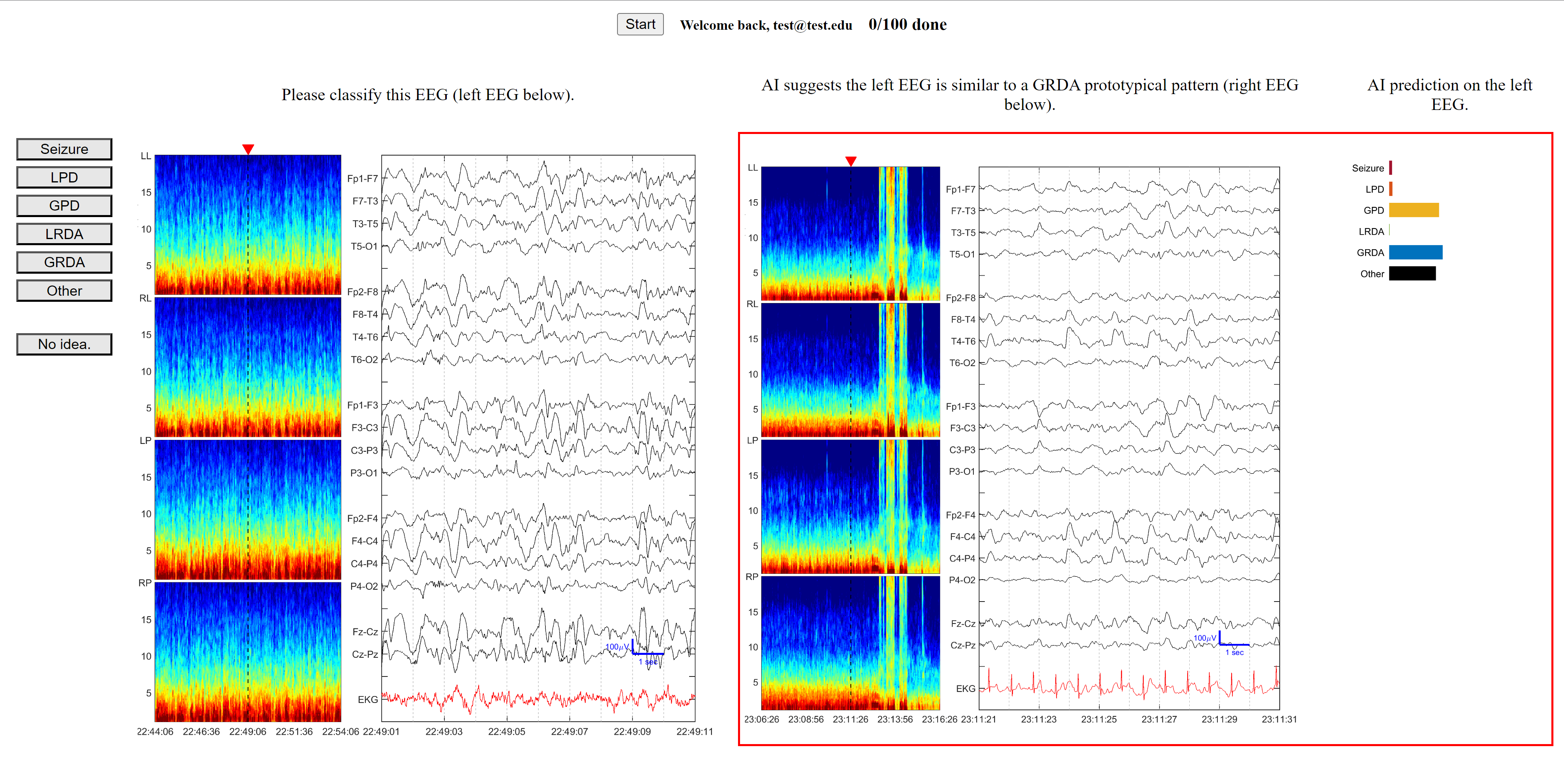}
  \caption{The user study interface when AI assistance is provided. From left to right: buttons to select the EEG pattern category; the EEG sample to be categorized; a comparable EEG prototype provided by the model; a bar chart of class predictions from the model.}
  \label{fig:userstudy_screenshot}
\end{figure}

Each participant was provided with basic training materials for identifying the five IIIC EEG patterns. A special user interface was designed for this user study, shown in Figure \ref{fig:userstudy_screenshot}. The study contains two stages; in both stages, users were asked to classify the same 100 samples as one of the six classes (one of five IIIC patterns or "Other") or choose "No idea." The two stages were set two weeks apart. Users were randomly split into two groups. One group was given AI assistance only in the first stage, and the other group was given AI assistance only in the second stage. The 100 samples were selected from the test set such that the classes were equally represented, and that the expert annotators had a high level of agreement with each other, and to maximize the patient diversity such that a patient does not appear twice within the same class. After the users completed both stages, they were asked to complete a survey with questions regarding the study.

\subsection*{External Validation} 
To validate model performance across institutions, we collected a new dataset of 1,500 events from 327 ICU patients who underwent continuous EEG as part of clinical care at Brigham and Women's Hospital. As in the original dataset, EEG experts ($n=10$) labeled the center 10-second of 50-second EEG segments and raters produced one of the six labels. 

\section*{Results \textcolor{red}{\sout{(618/3000 words)}}}

\subsection*{Model Classification Performance} 
\begin{figure}
     \centering
     \begin{subfigure}[b]{\textwidth}
         \includegraphics[width=\linewidth]{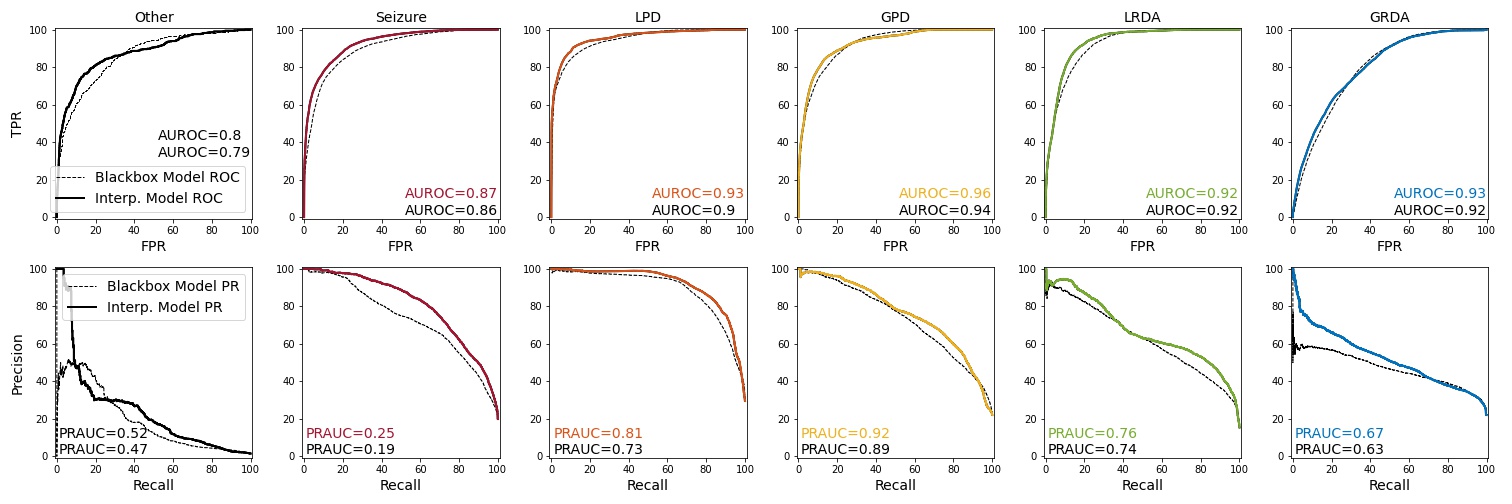}
        \caption{The receiver operating characteristic curves and precision-recall curves for ProtoPMed-EEG (solid lines) compared to SPaRCNet (dashed lines). ProtoPMed-EEG has statistically significantly higher AUROC and AUPRC.}
        \label{fig:results}
    \end{subfigure}
     \hfill
     \newline
     \begin{subfigure}[b]{0.45\textwidth}
         \centering
         \includegraphics[width=\textwidth]{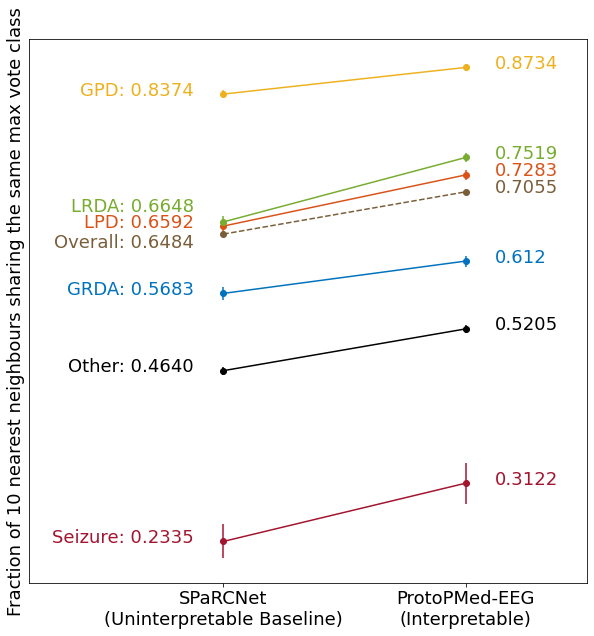}
         \caption{Neighborhood analysis by max}
         \label{fig:slope_charts_na_max}
     \end{subfigure}
     \hfill
     \begin{subfigure}[b]{0.45\textwidth}
         \centering
         \includegraphics[width=\textwidth]{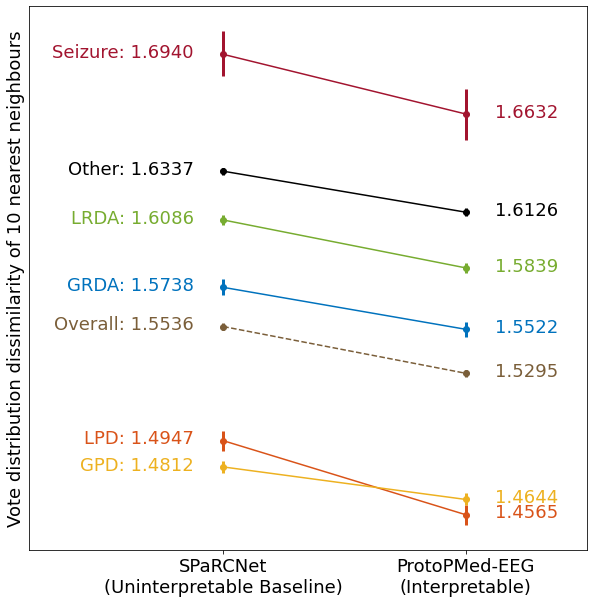}
         \caption{Neighborhood analysis by annotator vote distribution (lower values mean a more consistent neighborhood)}
         \label{fig:slope_charts_na_vote}
     \end{subfigure}
     \caption{Performance Evaluation: We compare the \hyperref[fig:results]{(a)} AUROC scores, AUPRC scores, \hyperref[fig:slope_charts_na_max]{(b)} neighborhood analysis by max vote \textcolor{red}{(higher value indicates better clustering by class, uses only majority vote of each sample)} and \hyperref[fig:slope_charts_na_vote]{(c)} neighborhood analysis by annotator vote distributions \textcolor{red}{(lower value indicates better clustering by class)} between the uninterpretable SPaRCNet \cite{older_eeg_paper} and our interpretable ProtoPMed-EEG. \textcolor{red}{\sout{The uncertainty bounds for each point are smaller than the point itself.}} }
     \label{fig:slope_charts}
\end{figure}

We evaluate model performance using area under receiver operating characteristic curve (AUROC) scores, and area under precision-recall curve (AUPRC) scores. \textcolor{red}{Both AUROC and AUPRC are calculated using the predicted class probability output from the softmax layer of the model.} The classification performance of our interpretable model ProtoPMed-EEG statistically significantly exceeds that of the SPaRCNet \cite{older_eeg_paper} in distinguishing Seizures, LPDs, GPDs, LRDAs, and GRDAs, as measured both by AUROC and AUPRC scores (p<0.001). Results for ROC and PR curve analysis are shown in Figure \ref{fig:slope_charts}. Seizure vs no-seizure classification performance can be found in Figure \ref{fig:slope_charts}(a) under the heading ``Seizure.'' 
For comparing AUROC scores, we use the Delong test \cite{delong1988comparing} for statistical significance. For AUPRC comparisons, we test for statistical significance using the bootstrapping method with 1000 bootstrap samples. For more detail on these statistical significance tests, refer to Appendix \ref{app:sig}. These findings hold when bootstrapping by patient or by sample.

\subsection*{User Study} 
\begin{figure}
  \centering
    \begin{subfigure}[b]{0.49\textwidth}
         \centering
         \includegraphics[width=\textwidth]{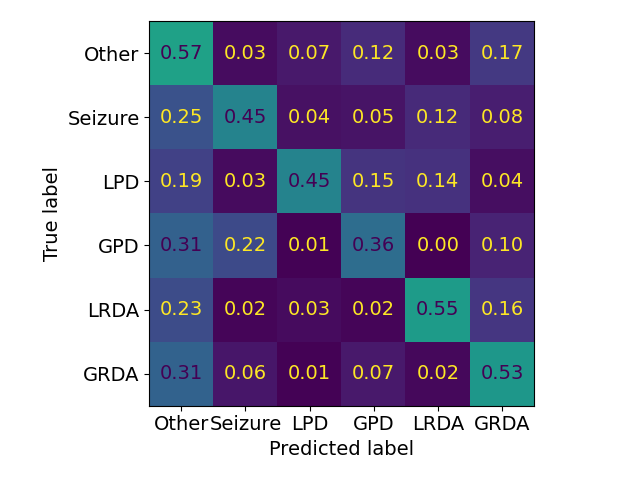}
         \caption{Average error matrix across all users without AI.}
    \end{subfigure}
    \hspace{-3em}
    \begin{subfigure}[b]{0.5\textwidth}
         \centering
         \includegraphics[width=\textwidth]{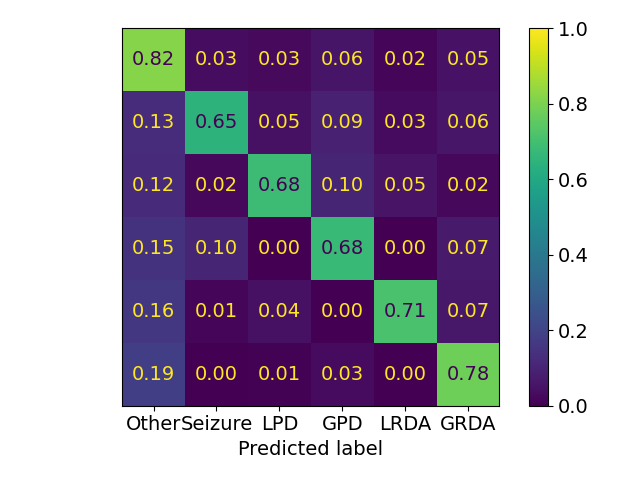}
         \caption{Average error matrix across all users with AI.}
    \end{subfigure}
    \newline
    \begin{subfigure}[b]{0.6\textwidth}
         \centering
         \includegraphics[width=\textwidth]{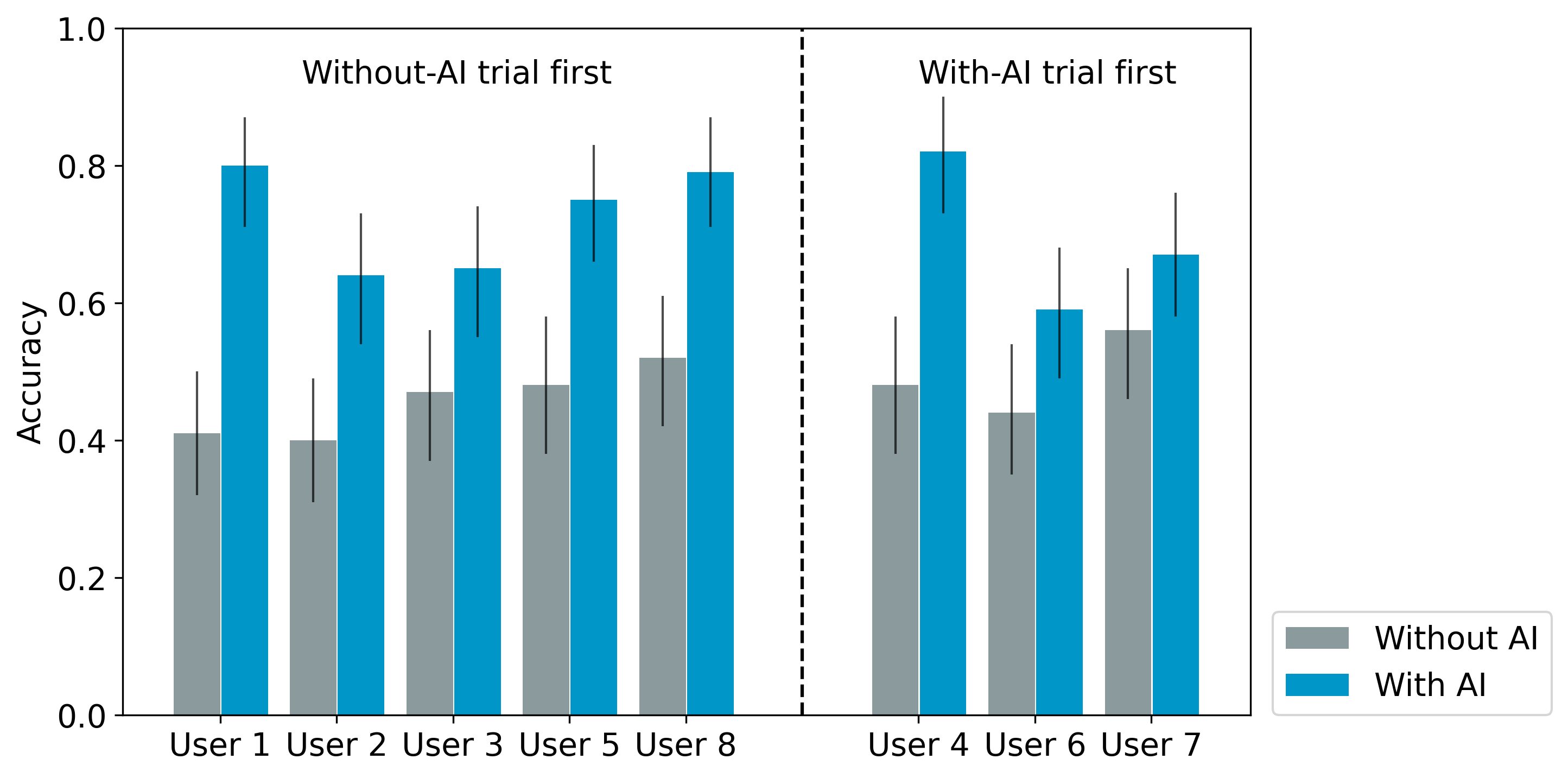}
         \caption{EEG pattern classification performance of the users with and without AI. All users performed significantly better ($p<0.05$) while provided with AI assistance.}
    \end{subfigure}
  \caption{The results of the user study.}
  \label{fig:user_study_acc}
\end{figure}

Of the 13 users invited, eight completed both stages of the study, two completed only the first stage but did not complete the second stage two weeks later, and seven filled out the post-study survey. Participant dropout rates are similar in both stages. None of the participants received performance feedback. Mean user \textcolor{red}{accuracy in identifying the correct class} improved from 47\% to 71\% when using AI compared to without. Using a one-sided student t-test, we found that all users statistically significantly improved with $p<0.05$ as shown in Table \ref{tab:user_study_sigs}. Refer to Figure \ref{fig:user_study_acc} for individual results. The users took longer to decide when presented with AI assistance and the decision was more likely to be correct. The average time taken to make a decision was 32 ± 33 seconds with AI assistance compared to 25 ± 39 seconds without. We observe an increase in mean inter-rater reliability (IRR). Rater-to-majority IRR has percentage agreement of 90\% with AI assistance and 86\% without; Cohen’s kappa scores of 66\% with AI assistance and 48\% without. Mean rater-to-rater IRR has percentage agreement of 62\% with AI assistance and 47\% without; Cohen’s kappa scores of 54\% with AI assistance and 35\% without.


In the post-study survey, all seven users indicated that they felt their ability to identify EEG patterns improved after completing the stage with AI, while only three users felt they improved after the stage without AI. All seven users would recommend this system to medical professionals learning to identify these patterns. Further analysis of the user study results can be found in Appendix \ref{app:user_study}.

\subsection*{External Validation} 
We calculate the model-rater agreement using AUROC and AUPRC. On the external dataset, we find an average AUROC of 0.85 and an average AUPRC of 0.61. On the internal dataset, we find an average AUROC of 0.91 and an average AUPRC of 0.74. Our model maintained high predictive performance, despite shifts in class distribution and annotator population. There is a large shift in class distribution between the two datasets, with the ``Other'' class making up 60\% of the external dataset, but only 22\% of the internal dataset. There is also a shift in the population of annotators between the internal and external datasets, with 10 annotators for the external dataset and 124 for the internal dataset. Further details can be found in Appendix \ref{app:ext_val}.

\subsection*{Model Interpretability Quantitative \& Qualitative Performance Evaluation} 
A key part of our explanation is to provide prototypical EEG samples that are comparable to the sample being analyzed. The interpretable model judges two samples' similarity from their distance in the latent space (i.e., Figure \ref{fig:3modes_of_interp}(a).) As the explanation depends on the meaningful placement of samples in the latent space, we expect the feature extractor to place samples with high resemblance close together. In an ideally structured latent space, we have neighborhoods of consistent samples -- in particular, they share the same class label and other medically relevant characteristics. To evaluate the explanatory power of the learned prototypes and the meaningfulness of the latent space, we designed two metrics to measure neighborhood consistency. In Figure \ref{fig:slope_charts_na_max}, for each sample in the test set, we calculate the fraction of the 10 nearest test set neighbors where the class with the most votes is the same as for the sample  (``by max''). For Figure \ref{fig:slope_charts_na_vote}, we consider the neighborhood analyses ``by vote,'' where for each sample, we calculate the mean cross-entropy of the vote distribution of the sample with the vote distribution of each of the 10 nearest neighbors. Here, we consider cross entropy as a discrete distribution across classes and check whether the cross entropy of the test point matches the distribution of classes from the nearest neighbors. The interpretable model performs statistically significantly better than the black box across all metrics and classes with $p<0.05$ for each comparison. Qualitative assessments of the neighborhoods are explored in Appendix \ref{app:neighbor_analyses}.

\subsection*{Mapping the Ictal-Interictal-Injury Continuum}
In Figure \ref{fig:3modes_of_interp}c, the coloring and distance between samples (points) are based on the model class scores for each sample. This results in a structure with outer points (arms) corresponding to single classes and reveals dense, thread-like paths mapping a gradual change between IIIC classes. This provides evidence for the concept of a ``continuum'' between ictal and interictal EEG patterns, as a set of well-separated classes would instead be represented by disconnected islands. This morphology is consistent across models initiated with different random seeds. This perspective is supported by the correlation between our model's predictions and the labeling experts' opinions. Our model successfully identifies samples that are categorized as ``between classes,'' where the class probabilities assigned by the model closely match the distribution of expert votes (i.e. split across two or more classes).

We further sampled along paths between each pair of IIIC patterns, and produced videos demonstrating the smooth continuum from one pattern to the other. Videos are provided at \url{https://warpwire.duke.edu/w/8zoHAA/}.

\section*{Discussion \textcolor{red}{\sout{(802/3000 words)}}}

In this study, we developed the first inherently interpretable deep learning model to classify IIIC activity. This study includes a user study with first-of-its-kind AI assistance using case-based explanations. We showed that when users are provided with interpretable AI assistance, their accuracy in predicting IIIC EEG patterns significantly improves, demonstrating the efficacy of this system for human-AI collaboration. We showed that the model generalized to a dataset from another hospital. Compared to the current state-of-the-art black box model for this task, the interpretable model achieved better neighborhood analysis scores, indicating that it learned \textit{purer neighborhoods in the latent space.}  That is, the geometry of our latent space, which groups samples from the same class together, forms neighborhoods without many ``other-class'' samples. This is useful for providing related EEG samples as part of its explanations. Our work thus yielded advances in both the model's predictive performance and interpretability. 

Machine learning, and specifically deep learning, has been used for EEG classification tasks including seizure detection with satisfactory predictive performance. Previous studies have produced fully automated black box models \cite{older_eeg_paper,tzallas2012automated,craik2019deep}, and black box models with \textit{post hoc} explanations \cite{gradcameegseizure1, gradcameegseizure2, filtervizLRP, shapvalues, featuremappermutationentropy}  in an effort to address interpretability challenges. However, at their best, \textit{post hoc} explanation methods only approximate model reasoning, and different methods will generate conflicting \textcolor{red}{\sout{reasoning} explanations}. In many cases, it is guaranteed that the explanation will not match true reasoning. In contrast, our explanation follows the exact path within the model as the prediction generation, thus offering \textit{perfect explanation faithfulness}. Our interpretable model goes beyond automated detection in black box models, providing clinicians with the means to validate diagnoses. The explanations produced by our model include a graphical representation of the sample's relative position to all learned prototypical samples along the underlying ictal-interictal-injury continuum, visual comparisons to relevant samples, and an easy-to-understand scoring system based on the learned similarity to the prototypical samples. These explanation components help users gauge the appropriate level of trust for a specific prediction based on the model's explicit reasoning. 

Our model is not only novel in its applications to neurology but also provides substantial improvements to the existing interpretable prototype-based neural network literature. While there exist interpretable deep learning models for medical applications, most are limited to the computer vision domain. \citet{barnett2021case} provides an inherently interpretable system for a breast mass classification task, but is limited to computer vision applications using prototypes that represent one part of an image. In past work on leveraging prototypes to provide explanations for model predictions \cite{zhang2020tapnet,huang2019deep,gee2019explaining}, each prototype was limited to represent a single class; this setting is insufficient for mapping IIIC EEG signals as some present defining features of two classes. Our introduction of \textit{dual-class} prototypes enables our model to place prototypes between two classes in the latent space, providing insights into EEG patterns in the transitional states. 

The ProtoPMed-EEG model's expert-level predictive performance and its interpretable nature make it a promising candidate for application in clinical ICU settings. The enhancements showcased by participants in the user study, where a purpose-built graphical user interface incorporates the model's explanations, highlight its potential to mitigate human subjectivity and enhance user classification accuracy, particularly for challenging IIIC EEG patterns. This system has the potential to reduce reliance on human experts for EEG categorization and reading tasks in clinical ICU settings; in cases where expert-level supervision is not available, physicians without fellowship training in (IIIC) EEG readings engaged in providing patient care could utilize this system to reach diagnoses with reduced subjectivity, averting potential misinterpretations and ensuring comprehensive patient care. The implementation of this system could improve patient care in regions with limited access to optimal medical personnel resources. In addition, with little to no modifications, this system presents a low-cost and interactive training tool for physicians and clinical practitioners on IIIC EEG pattern recognition and classification. Beyond the above-mentioned values, the inherent interpretability of our model would facilitate adoption of this deep learning model in real-world practice, as it provides humans with adequate visibility into the model's reasoning process to reduce potential misdiagnoses. 

There are limitations to our study. \textcolor{red}{Firstly, we use the majority vote of the 124 raters as ground truth. Even though the rater population is large, it is still possible that a different group of raters may yield different ground truth labels. Secondly, t}he interpretability is limited to the final steps of the model, so a clinician must still infer how the relevant qualities (i.e., peak-to-peak distance, amplitude, burst suppression ratio) are weighed in the model's assessment of similarity. In cases where this is not clear, qualitative neighborhood analyses are available for examination. Future work could account for known qualities of interest explicitly. Despite this limitation, the interpretability provided by this model greatly surpasses that of the state-of-the-art for this task.

In conclusion, we developed an interpretable deep learning algorithm that accurately classifies six clinically relevant EEG patterns, and offers faithful explanations for its classifications by leveraging prototype learning. We show that users have superior classification performance when provided with this AI assistance. The comprehensive explanations and the graphical user interface enable follow-up user studies working toward clinical applications, including diagnostic assistance and education.

\section*{Acknowledgments}
We acknowledge support from the National Science Foundation under grants IIS-2147061 (with Amazon), HRD-2222336, IIS-2130250, and 2014431, from the NIH (R01NS102190, R01NS102574, R01NS107291, RF1AG064312, RF1NS120947, R01AG073410, R01HL161253). \textcolor{red}{O.T. received support from the NIH P20GM130447 Cognitive Neuroscience and Development of Aging (CONDA) Award and the DHHS LB606 Nebraska Stem Cell Grant.}

We also acknowledge Drs. Aaron F. Struck, Safoora Fatima, Aline Herlopian, Ioannis Karakis, Jonathan J. Halford, Marcus Ng, Emily L. Johnson, Brian Appavu, Rani A. Sarkis, Gamaleldin Osman, Peter W. Kaplan, Monica B. Dhakar, Lakshman Arcot Jayagopal, Zubeda Sheikh, Olha Taraschenko, Sarah Schmitt, Hiba A. Haider, Jennifer A. Kim, Christa B. Swisher, Nicolas Gaspard, Mackenzie C. Cervenka, Andres Rodriguez, Jong Woo Lee, Mohammad Tabaeizadeh, Emily J. Gilmore, Kristy Nordstrom, Ji Yeoun Yoo, Manisha Holmes, Susan T. Herman, Jennifer A. Williams, Jay Pathmanathan, Fábio A. Nascimento, Mouhsin M. Shafi, Sydney S. Cash, Daniel B. Hoch, Andrew J. Cole, Eric S. Rosenthal, Sahar F. Zafar, and Jimeng Sun, who played major roles in creating the labeled EEG dataset and SPaRCNet used in the study.

We would also like to acknowledge Rachel Choi, Justin Sattin, Kayla N Haffley, Anthony P Tran, Ushna Khan, Luana Rodrigues Dos Santos, HyoJin Park, and Emma Locke for participating in our user study.

\textcolor{red}{We thank Piotr Suder for lending his statistical expertise.}

\section*{Conflict of Interest Disclosures}
Dr. Westover is a co-founder of Beacon Biosignals, which played no role in this work.

\section*{Author Contribution Statement}
Idea conception and development: AJB, ZG, JJ, BW, CR. Interpretable model code: ZG, AJB. GUI code: JJ. User study interface code: JJ. Data preparation: JJ, ZG, AJB, WG. Writing: AJB, ZG, JJ, BW, CR. \textcolor{red}{External validation data annotation: PK, WYK, IK, AH, LAJ, OT, OS, GO, DG.}


\FloatBarrier
\bibliographystyle{unsrtnat}
\bibliography{main}

\newpage
\appendix
\FloatBarrier

\FloatBarrier
\setcounter{figure}{0}

\section{Model Details}\label{app:model_details}
\FloatBarrier

\begin{figure}[ht]
    \centering
    \includegraphics[scale=0.57]{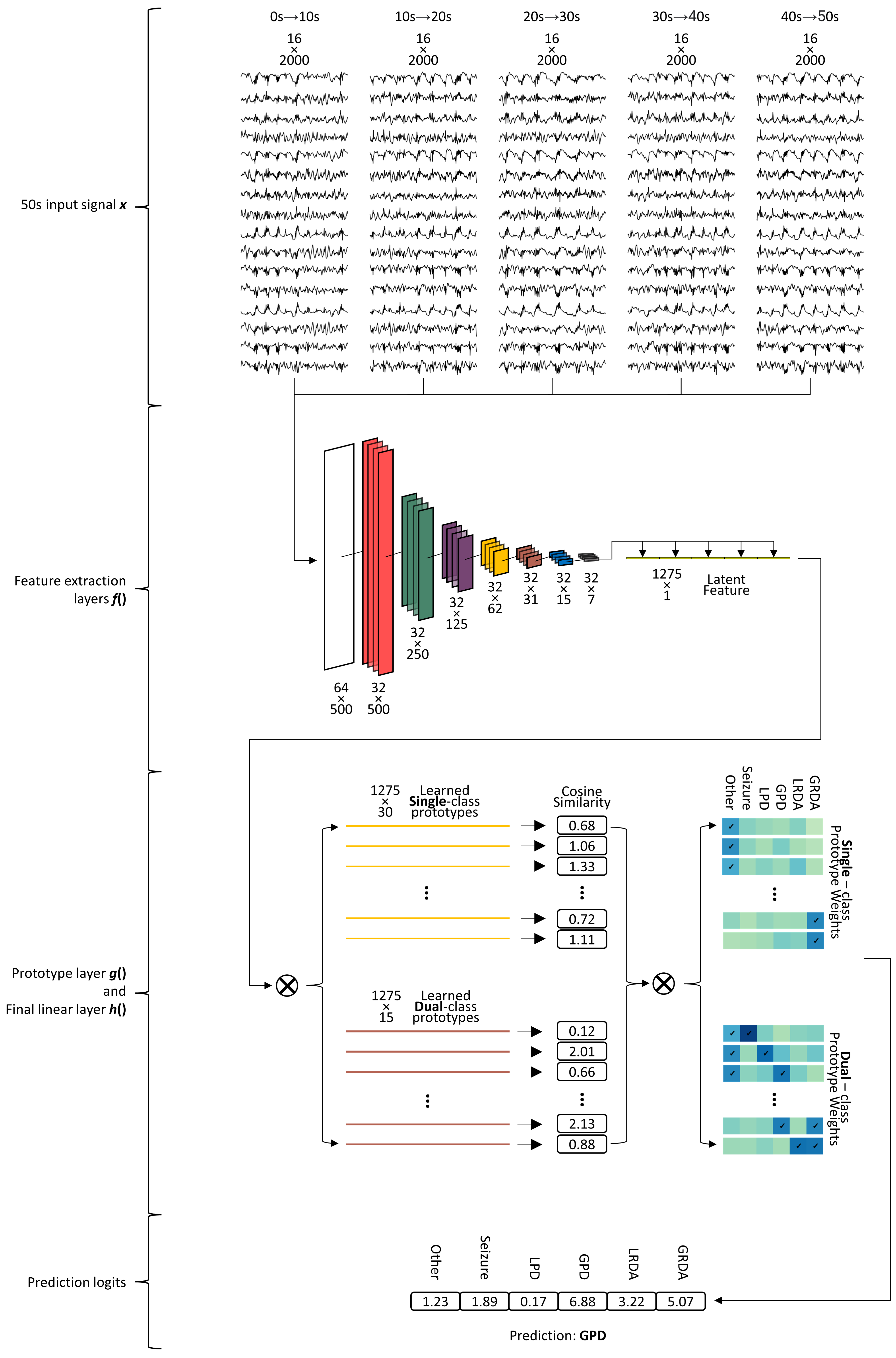}
    \caption{This is a detailed architecture plot of the ProtoPMed-EEG, indicating output dimensions of key components of the model. The check marks ($\checkmark$) in the figure indicate the class connection of each prototype, each single-class prototype has one check mark, and each dual-class prototype has two check marks. The colored heatmap represents class affinity values.}
    \label{fig:detailed_arch}
\end{figure}

\subsection{Model Architecture}

Let $D=\{\mathbf{x}_i, y_i\}_{i=0}^n$ be our dataset of $n$ 50-second EEG samples $\mathbf{x}$ with class labels $y$ indicating the EEG pattern of Seizure, LPD, GPD, LRDA, GRDA, or Other. Refer to Figure \ref{fig:detailed_arch} for a diagram of the model architecture. Our model consists of feature extraction layers $f_{\omega_f}()$, prototype layer $g_{\omega_g}()$ and final linear layer $h_{\omega_h}()$. For the $i^{\text{th}}$ EEG sample, corresponding model prediction $\hat{y}_i$ is calculated as 
\begin{equation}
    \hat{y}_i=h_{\omega_h}(g_{\omega_g}(f_{\omega_f}(\mathbf{x}_i))).
\end{equation}

The feature extraction layers $f_{\omega_f}()$ consist of all but the final layer of the neural network architecture from \citet{older_eeg_paper}, which was trained on the same dataset. The model architecture from \citet{older_eeg_paper} was designed to handle 10-second samples, where (when removing the final layer) the output for a single 10-second EEG sample is a vector of length 255. Our model uses 50-second samples by concatenating the five 10-second outputs into a single 1275-length vector which we will denote as $f_{\omega_f}(\mathbf{x})$. 

The input to prototype layer $g_{\omega_g}()$ is the output from $f_{\omega_f}(\mathbf{x})$. The set of $m$ prototypes is $P=\{p_j\}_{j=0}^m$. Each prototype is a vector of length 1275 that exists in the same \textit{1275-dimensional latent space} as outputs $f_{\omega_f}(\mathbf{x})$. For each prototype, the similarity score $s$ between test sample $\mathbf{x}$ and prototype $p_j$ is 
\begin{equation}
    s(f_{\omega_f}(\mathbf{x}), p_j) = \left(\frac{a f_{\omega_f}(\mathbf{x})}{\|f_{\omega_f}(\mathbf{x})\|}\right) \cdot \left(\frac{p_j}{\|p_j\|}\right),
\end{equation}
where $a=64$ as in \citet{deng2019arcface}. This similarity score is 0 for orthogonal vectors and 64 for identical vectors.

The prototypes are each given a class identity by final linear layer $h_{\omega_h}()$. $h_{\omega_h}()$ has size $m \times C$, where $m$ is the number of prototypes and $C$ is the number of classes. We can think of this as a $C$-length vector for each prototype, which determines how similarity to the prototype affects class score. For example, a single-class class-two prototype has a corresponding $C$-length vector in $h_{\omega_h}()$ that is initialized as $(-1, -1, 1, -1, -1, -1)$. This means that if a sample is similar to this prototype, the score for class two will increase and the score for all other classes will decrease. (Note that the class indices range from 0 to the number of classes, which is why class 2 is represented by the third index of the vector.) We call this the \textit{class connection vector} for the prototype. 

As mentioned, we introduce the idea of dual-class prototypes in this paper. For example, a prototype of both class zero and class five would have class connection vector initialized as $(1, -1, -1, -1, -1, 1)$. Though prototypes shared between classes were used in previous work by \citet{rymarczyk2021protopshare}, their method differs in that it merges similar prototypes, whereas our method learns prototypes that are explicitly in-between classes and do not fit into either class individually.

The number of prototypes is selected as a hyperparameter before training, though it is possible to prune prototypes \textit{post hoc}. If there are too few prototypes, the similarity between a test sample and the displayed prototype will not be clear to the end user. We need enough prototypes to cover the space so that the similarity between any test sample and the nearest prototype is visually clear. If there are too many prototypes, the model overfits and the latent space has less structure, impairing interpretability (however, we can detect overfitting using a validation set, as usual). We may also end up with duplicate prototypes when there are too many prototypes. For this domain, we decided on 45 prototypes consisting of five single-class prototypes for each class (30 total) and one dual-class prototype for each pair between two classes (15 total, 6 choose 2).

\subsection{Model Training}
The feature extraction layers $f_{\omega_f}()$ are initialized with pretrained weights from the uninterpretable SPaRCNet of \citet{older_eeg_paper}.

Our model is trained in four stages: (1) warm up, (2) joint training, (3) projection of the prototypes and (4) last layer optimization. Training starts with the warm up stage for 10 epochs. After warm up, the training cycles from joint training for 10 epochs, to the projection step, to last layer optimization for 10 epochs, before returning to joint training for the next cycle. We continue for 80 total epochs, as training typically converges between 30 and 40 epochs.

\textbf{(1) Warm up.} During the warm up stage, only the prototype layer weights in $g_{\omega_g}()$ change. We use a learning rate of 0.002. The objective function for the warm up stage is: 
\begin{equation}
    \min_{\omega_g} \left(
    \mathrm{CrsEnt} + \lambda_c ~ \ell_\mathrm{clst} + \lambda_s ~ \ell_\mathrm{sep} + \lambda_o ~ \ell_\mathrm{ortho}\right), 
\end{equation}
where $\mathrm{CrsEnt}$ is the cross entropy, $\ell_\mathrm{clst}$ and $\ell_\mathrm{sep}$ encourage clustering around meaningful prototypes in the latent space and $\ell_\mathrm{ortho}$ encourages orthogonality between prototypes. The weights on each loss term are as follows: $\lambda_c=-0.8$, $\lambda_s=-0.08$, and $\lambda_o=100$. Adapted from \citet{chen2019this} and \citet{barnett2021case} to use cosine similarity instead of L2 similarity,
\begin{equation}
    \ell_\mathrm{clst} = \frac{1}{n}\sum_{i=1}^{n} \min_{j:\text{class}(\mathbf{p}_j)=y_i}  s(f_{\omega_f}(\mathbf{x}_i), p_j) ,\hspace{0.5cm}\ell_\mathrm{sep} = -\frac{1}{n}\sum_{i=1}^{n} \min_{j:\text{class}(\mathbf{p}_j) \neq y_i}  s(f_{\omega_f}(\mathbf{x}_i), p_j) \label{eq:clst_sep_costs}.
\end{equation}
$\ell_\mathrm{clst}$ encourages each training sample to be close to a prototype of its own class, and $\ell_\mathrm{sep}$ encourages each training sample to be far from prototypes not of its own class. $\ell_\mathrm{ortho}$ is defined as
\begin{align}
    \label{eq:orthogonality_loss}
    \ell_\mathrm{ortho} =& \frac{1}{a^2}\left( \sum_{j=0}^m \sum_{j'=0}^m (s(p_j, p_{j'}))^2 - \sum_{j=0}^m (s(p_j, p_j))^2 \right)\text{, or equivalently} \\
    \ell_\mathrm{ortho} =& \| \mathbf{P}\mathbf{P}^T - I \|_F^2,
\end{align}

where $\mathbf{P}$ is the $m$ $\times$ $1275$ matrix of normalized prototype vectors and $\|\|_F^2$ is the squared Frobenius norm.

\textbf{(2) Joint training.} During joint training, weights from $f_{\omega_f}()$, $g_{\omega_g}()$ and $h_{\omega_h}()$ are all trained simultaneously. We chose a learning rate of 0.0002 for training the weights in $f_{\omega_f}()$, 0.003 for the weights in $g_{\omega_g}()$ and 0.001 for the weights in $h_{\omega_h}()$.  The objective function is: 
\begin{equation} \label{eq:joint_objective}
    \min_{\omega_f, \omega_g, \omega_h} \left(\mathrm{CrsEnt} + \lambda_c ~ \ell_\mathrm{clst} + \lambda_s ~ \ell_\mathrm{sep} + \lambda_o ~ \ell_\mathrm{ortho} + \lambda_l ~ \ell_\mathrm{lll1}\right),
\end{equation}
where $\mathrm{CrsEnt}$, $\ell_\mathrm{clst}$, $\ell_\mathrm{sep}$, $\ell_\mathrm{ortho}$, $\lambda_c$, $\lambda_s$, and $\lambda_o$ are as above, while $\ell_\mathrm{lll1}$ is the L1 loss on the weights in the last layer and $\lambda_l=0.0001$.

\textbf{(3) Projection step.} During the projection step, we project the prototype vectors on the nearest (most similar) EEG samples from the projection set. The projection set $D'$ is the subset of the training set $D$ that has at least 20 expert annotations of the class. This allows us to visualize a prototype as, after projection, the prototype exactly equals the latent space representation of the projection set example. This projection step is
\begin{equation}\label{eq:projection}
    p_j^{\text{updated}} = f_{\omega_f}\left( \underset{\mathbf{x} \in D'}{\text{argmax}} \ s\left(f_{\omega_f}(\mathbf{x}), p_j\right) \right).
\end{equation}

\textbf{(4) Last layer optimization.} During last layer optimization, weights from $f_{\omega_f}()$ and $g_{\omega_g}()$ are frozen and only weights from $h_{\omega_h}()$ may change. The learning rate is 0.001 and the objective function is: 
\begin{equation} \label{eq:lastlayer_objective}
    \min_{\omega_h} \left( \mathrm{CrsEnt} + \lambda_l ~ \ell_\mathrm{lll1}\right),
\end{equation}
where $\mathrm{CrsEnt}$, $\ell_\mathrm{lll1}$, $\lambda_l$ are as above.\\

Note: We considered using the margin loss from \citet{donnelly2022deformable}, which encourages wide separation in the latent space between classes. We noted much worse results when margin loss was included. We speculate that this is because the classes are not clearly separated in this domain, but instead exist on a continuum. With margin loss included, the network is penalized for placing an EEG sample between two classes, yet we know that there exist dual-class samples (i.e., SZ-GPD). This could account for the drop in accuracy when margin loss is introduced.

The model is trained with an objective function containing loss terms to encourage accuracy, a clustering structure in the latent space, and separation between prototypes. Training is completed in 4 hours using two Nvidia V100 GPUs.

\FloatBarrier

\section{Model Performance Comparison}
See Table \ref{tab:aucs}.
\begin{table*}[h!]
  \caption{AUROC, AUPRC and neighborhood analysis of the interpretable model compared to uninterpretable state-of-the-art SPaRCNet \cite{older_eeg_paper}. Each prediction problem is one-vs-all. The column name ``All'' refers to a mean for all classes weighted by the number of samples in each class. 
  95\% confidence intervals are shown in square brackets. 
  We use the bootstrapping method described in Appendix \ref{app:boot} for AUPRC, AUROC and the neighborhood analyses. The test set size $N$ is $35740$ cEEG samples. Our results show statistically significant improvements over SPaRCNet for all comparisons, see Appendix Table \ref{tab:significance_test_percent} for AUROC and AUPRC significance test results and Appendix Table \ref{tab:significance_test} for neighborhood analysis significance test results.}
  \label{tab:aucs}
  \scriptsize
  \begin{tabular}{llccccccc}
    \toprule
    & & Other & Seizure & LPD & GPD & LRDA & GRDA & All\\
    \midrule
    \multirow{2}{1cm}{AUROC} & Interp. & \textbf{0.80 [0.79, 0.80]}& \textbf{0.87 [0.86, 0.89]}& \textbf{0.93 [0.92, 0.93]}& \textbf{0.96 [0.95, 0.96]}& \textbf{0.92 [0.92, 0.93]}  & \textbf{0.93 [0.93, 0.94]}& \textbf{0.91 [0.90, 0.91]}\\
    & Uninterp. & 0.79 [0.79, 0.80]& 0.86 [0.85, 0.88]& 0.90 [0.90, 0.90] & 0.94 [0.94, 0.95]& 0.92 [0.92, 0.92]  & 0.92 [0.91, 0.92]& 0.89 [0.89, 0.90]\\
    \multirow{2}{1cm}{AUPRC} & Interp. & \textbf{0.52 [0.51, 0.54]}& \textbf{0.25 [0.21, 0.28]}& \textbf{0.81 [0.80, 0.82]}& \textbf{0.92 [0.91, 0.92]}& \textbf{0.76 [0.75, 0.77]}& \textbf{0.67 [0.65, 0.68]}& \textbf{0.74 [0.74, 0.75]}\\
    & Uninterp. & 0.47 [0.46, 0.48]& 0.19 [0.16, 0.23]& 0.73 [0.72, 0.74]& 0.89 [0.88, 0.89]&  0.74 [0.72, 0.75]& 0.63 [0.61, 0.64]& 0.70 [0.69, 0.70]\\

    Neighborhood & Interp. & \textbf{0.52 [0.51, 0.53]} & \textbf{0.31 [0.28, 0.34]}& \textbf{0.73 [0.72, 0.74]} & \textbf{0.87 [0.87, 0.88]} & \textbf{0.75 [0.75, 0.76]} & \textbf{0.61 [0.60, 0.62]} & \textbf{0.71 [0.70, 0.71]}\\
    Analysis by Max & Uninterp. & 0.46 [0
    46, 0.47]& 0.23 [0.21, 0.26]& 0.66 [0.65, 0.67]& 0.84 [0.83, 0.84] & 0.66 [0.66, 0.67]& 0.57 [0.56, 0.58]& 0.65 [0.65, 0.65] \\
    Neighborhood & Interp. &  \textbf{1.61 [1.61, 1.61]} &  \textbf{1.66 [1.65, 1.68]} &  \textbf{1.46 [1.45, 1.46]} &  \textbf{1.46 [1.46, 1.47]} &  \textbf{1.58 [1.58, 1.59]} &  \textbf{1.55 [1.55, 1.56]} &  \textbf{1.53 [1.53, 1.53]}\\
    Analysis by Votes & Uninterp. & 1.63 [1.63, 1.63] & 1.69 [1.68, 1.71] & 1.49 [1.49, 1.50]& 1.48 [1.48, 1.48] & 1.61 [1.61, 1.61]& 1.57 [1.57, 1.58]& 1.55 [1.55, 1.56]\\
    \bottomrule
  \end{tabular}

\end{table*}

\FloatBarrier

\section*{Graphical User Interface} \label{app:gui}
\begin{figure}
  \centering
     \includegraphics[width=\linewidth]{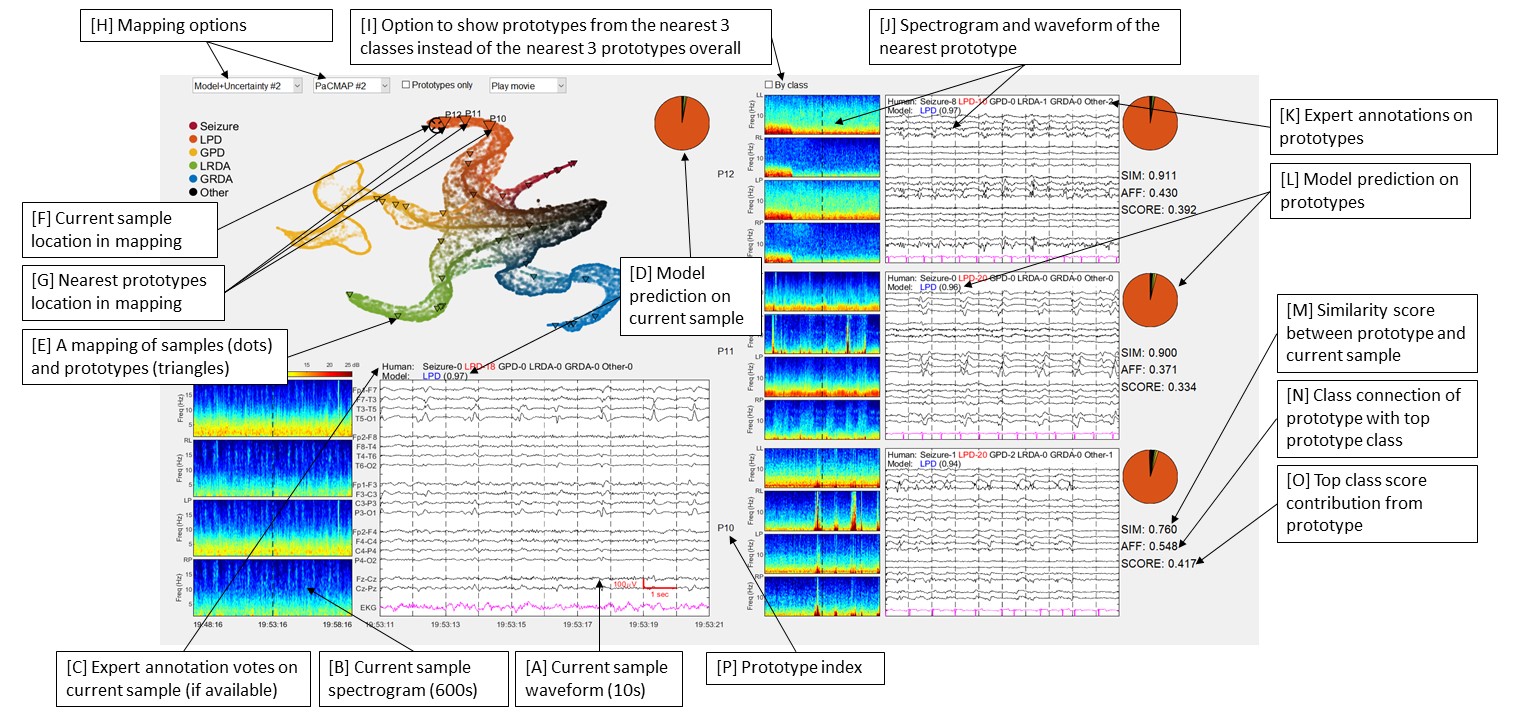}
     \caption{The graphical user interface (GUI) of the interpretable system. On the left top panel is the 2D embedding map, with each dot representing one EEG sample. Dots can be displayed with shading according to 9 different available schemes (human majority, model prediction, model uncertainty, Seizure burden, etc.). A user can click on the map to select any sample of interest; the 3 nearest prototypes are displayed on the right, ranked according to similarity score (SIM). For each sample/prototype, 10 seconds of the EEG and a 10-minute spectrogram (centered on the 10-second EEG segment) are displayed with human votes and model predictions shown on top of the EEG; a pie chart is provided to visualize the class distribution according to the model or human votes, depending on the selected color scheme. For each prototype, under its pie chart, we also list the values of three terms: similarity score (SIM), class connection (AFF), and class contribution score (SCORE).}
     \label{fig:GUI_screenshot}
\end{figure}

We also present a novel graphical user interface (GUI) that allows users to explore the interpretable model and its predictions on the test set. A screenshot is shown in Figure \ref{fig:GUI_screenshot}. 
On the bottom left of the GUI, we present the waveform [A], spectrogram [B], and the expert-annotator votes [C] of the currently selected sample. Below the expert-annotator votes, we display the model prediction [D] on the currently selected sample. On the upper left of the GUI, we present the 2D representation of the latent space that was generated using PaCMAP [E]. We can see the location of the current sample [F], as well as the location of the nearest prototypes [G] in the 2D representation. The color schema and mapping methods can be changed using drop-down menus [H]. Which prototypes are displayed can be changed using a selection box [I], where one option shows the nearest prototypes regardless of class and the other shows the nearest prototype from each of the three highest-score classes. Along the right-hand side, we display the waveform, spectrogram [J], expert-annotator votes [K] and model prediction [L] for the prototypical samples from the three displayed prototypes (in this case, the three nearest prototypes). For each of the displayed prototypes, we show the similarity score between the prototypical sample and the currently selected sample [M], the class-connection between the prototype and the predicted class (affinity) [N], and the class score added by that prototype to the predicted class [O].

\FloatBarrier
\section{Extended Analysis of the User Study}\label{app:user_study}
\FloatBarrier

\subsection{Study Design}
The same 100 samples were provided to all readers for both trials. The order of the samples was the same for all readers for the trial with AI, and was the same for all readers for the trial without AI; but these two orders are different. In other words, we shuffled the samples only as per task (but kept the same for all readers). We did so to avoid scenarios in which readers may memorize the question order. Both orders were random.
\subsection{User Information}
\label{sec:user_info}
All users were recruited by email. To preserve anonymity, Table \ref{tab:user_study_users} provides minimal information on the educational backgrounds of each participant who finished both stages of the user study. Participants included a registered nurse, an EEG technician, residents in neurology and dermatology, and specialists in stroke and neurophysiology.


\begin{table*}[ht]
  \centering
  \caption{User study participants.}
  \label{tab:user_study_users}
  \begin{tabular}{ll}
    \toprule
    User Number & Background \\
    \midrule
        1 & non-MD \\
        2 & non-MD \\
        3 & --- \\
        4 & MD \\
        5 & MD \\
        6 & MD \\
        7 & MD \\
        8 & MD \\
    \bottomrule
  \end{tabular}
\end{table*}








\subsection{Statistical Significance}
We performed a one-sided student t-test using the scikit learn function ttrel.

\begin{table*}[ht]
  \centering
  \caption{The statistical significance of the performance improvements of each user when AI assistance is provided. We report user accuracy in identifying the correct IIIC pattern and the 95\% confidence intervals follow in square brackets. Confidence intervals are calculated using the percentile method and 1000 bootstrap samples.}
  \label{tab:user_study_sigs}
  \begin{tabular}{rrrrrrr}
    \toprule
    User Number & Accuracy (no-AI) & Accuracy (with-AI) & $t$-statistic & $p$-value & Degrees of Freedom & Stage Order \\
    \midrule
        1 & 0.41 [0.32, 0.50] & \textbf{0.80 [0.71, 0.89]} &-7.36 & 2.73e-11 & 99 & no-AI first \\
        2 & 0.40 [0.31, 0.49] & \textbf{0.64 [0.54, 0.73]} &-4.34 & 1.70e-05 & 99 & no-AI first \\
        3 & 0.47 [0.37, 0.56] & \textbf{0.65 [0.55, 0.74]} & -3.60 & 2.50e-4 & 99 & no-AI first \\
        4 & 0.48 [0.38, 0.58] & \textbf{0.82 [0.73, 0.90]} & -6.84 & 3.27e-10 & 99 & with-AI first \\
        5 & 0.48 [0.38, 0.58] & \textbf{0.75 [0.66, 0.83]} & -5.52 & 1.38e-07 & 99 & no-AI first \\
        6 & 0.44 [0.35, 0.54] & \textbf{0.59 [0.49, 0.68]} & -2.89 & 0.00240 & 99 & with-AI first \\
        7 & 0.56 [0.46, 0.65] & \textbf{0.67 [0.58, 0.76]} & -1.74 & 0.0429 & 99 & with-AI first \\
        8 & 0.52 [0.42, 0.61] & \textbf{0.79 [0.71, 0.87]} & -5.10 & 8.07e-07 & 99 & no-AI first \\
    \bottomrule
  \end{tabular}
\end{table*}

\subsection{Qualitative Feedback}

Anonymized answers to the post-study qualitative survey can be provided on request by email to {\tt zhicheng.guo@duke.edu}.

Users described the system as:
\begin{itemize}
    \item "A useful reference when in trouble"
    \item "A useful tool [...] particularly when still learning"
    \item "It is useful to have the probabilities"
    \item "Much easier with the AI"
    \item "Helpful in identifying patterns"
    \item "A nice tool"
\end{itemize}

Area for improvement included:
\begin{itemize}
    \item "The role of the spectrograms wasn't clear"
    \item A user wanted to see a more diverse set of prototype EEGs
    \item "The AI seemed to struggle the most with coming up with comparable eeg."
\end{itemize}
\FloatBarrier
\subsection{User Study Error Matrices}
\begin{figure}
  \centering
    \vspace{-2em}
    \hspace{-2em}
    \begin{subfigure}[b]{0.29\textwidth}
         \includegraphics[width=\linewidth]{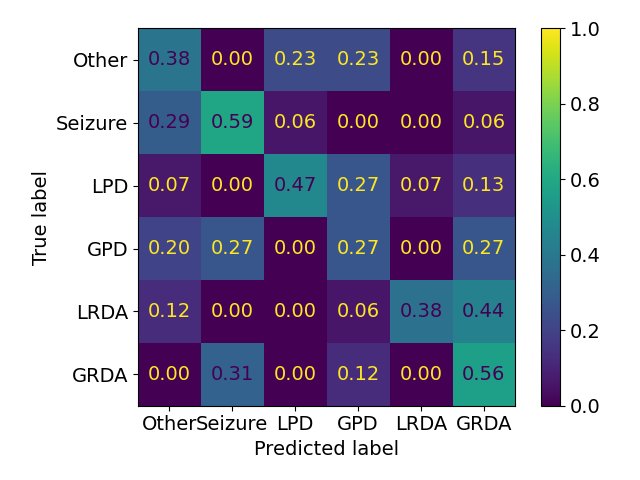}
        \caption{User 1 without AI.}
    \end{subfigure}
    \hspace{-1.3em}
    \begin{subfigure}[b]{0.29\textwidth}
         \centering
         \includegraphics[width=\textwidth]{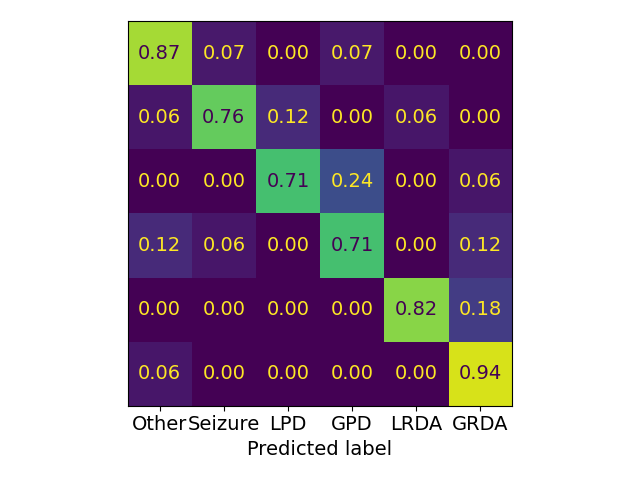}
         \caption{User 1 with AI.}
    \end{subfigure}
    \hspace{-3.35em}
    \begin{subfigure}[b]{0.29\textwidth}
         \centering
         \includegraphics[width=\textwidth]{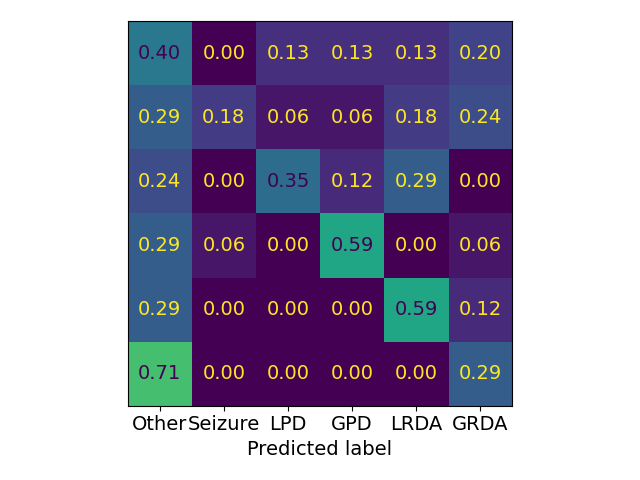}
         \caption{User 2 without AI.}
    \end{subfigure}
    \hspace{-3.35em}
    \begin{subfigure}[b]{0.29\textwidth}
         \centering
         \includegraphics[width=\textwidth]{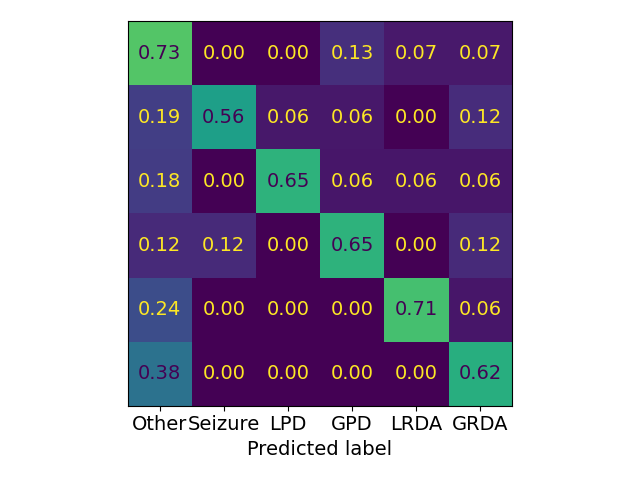}
         \caption{User 2 with AI.}
    \end{subfigure}
    \newline
    \begin{subfigure}[b]{0.29\textwidth}
         \includegraphics[width=\linewidth]{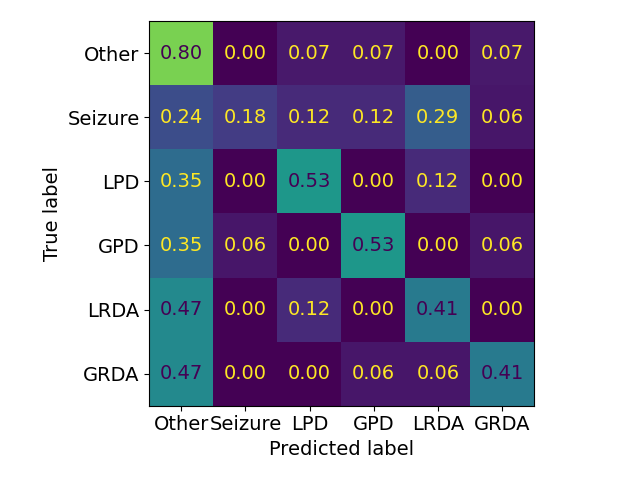}
        \caption{User 3 without AI.}
    \end{subfigure}
    \hspace{-2.8em}
    \begin{subfigure}[b]{0.29\textwidth}
         \centering
         \includegraphics[width=\textwidth]{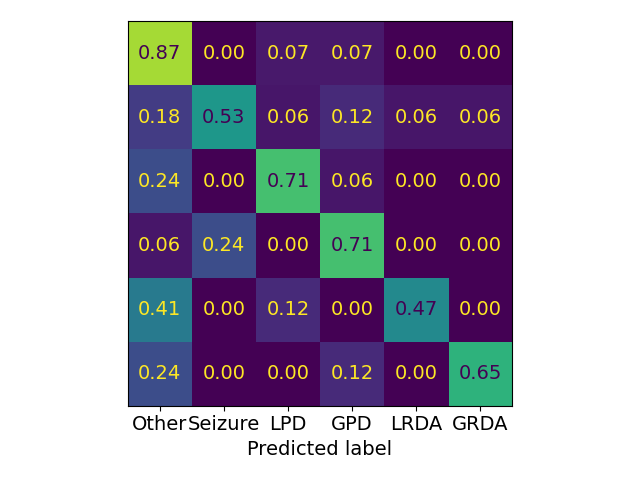}
         \caption{User 3 with AI.}
    \end{subfigure}
    \hspace{-3.35em}
    \begin{subfigure}[b]{0.29\textwidth}
         \centering
         \includegraphics[width=\textwidth]{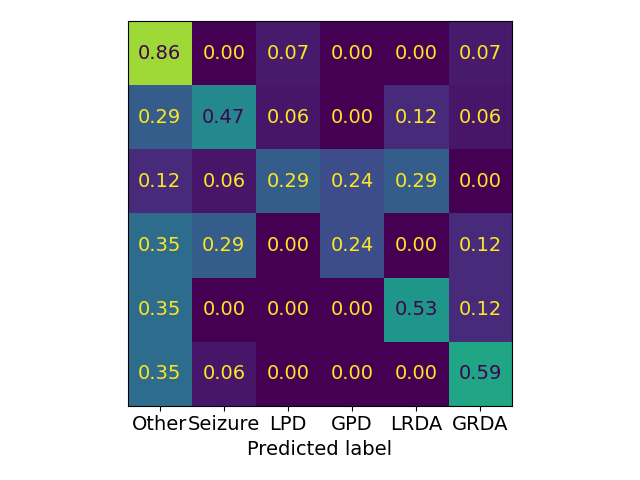}
         \caption{User 4 without AI.}
    \end{subfigure}
    \hspace{-3.35em}
    \begin{subfigure}[b]{0.29\textwidth}
         \centering
         \includegraphics[width=\textwidth]{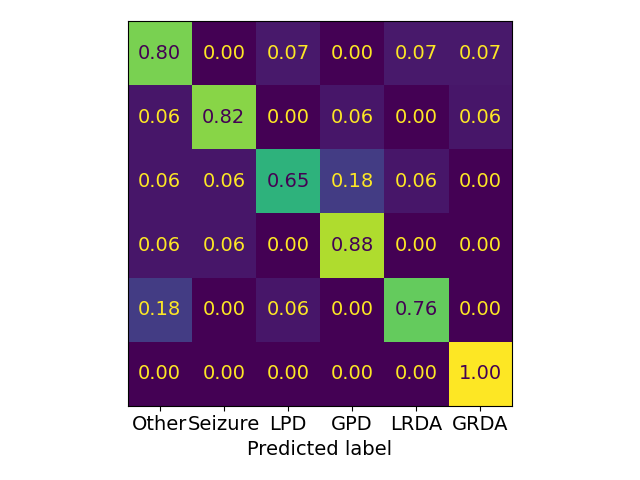}
         \caption{User 4 with AI.}
    \end{subfigure}
    \newline
    \begin{subfigure}[b]{0.29\textwidth}
         \includegraphics[width=\linewidth]{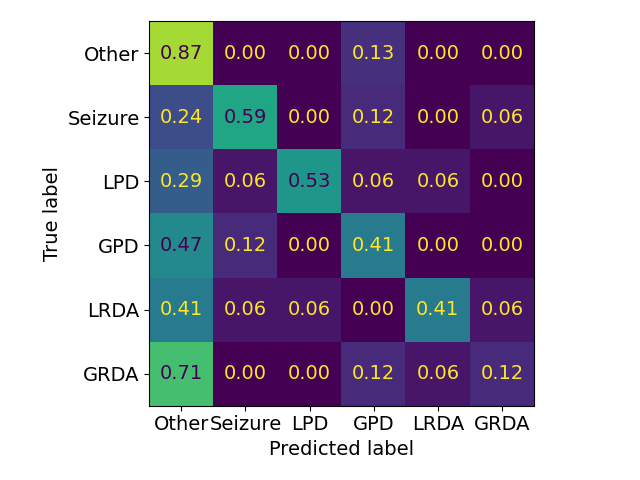}
        \caption{User 5 without AI.}
    \end{subfigure}
    \hspace{-2.8em}
    \begin{subfigure}[b]{0.29\textwidth}
         \centering
         \includegraphics[width=\textwidth]{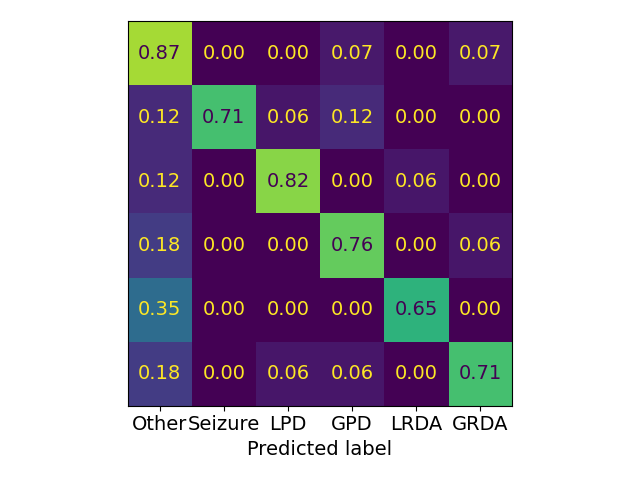}
         \caption{User 5 with AI.}
    \end{subfigure}
    \hspace{-3.35em}
    \begin{subfigure}[b]{0.29\textwidth}
         \centering
         \includegraphics[width=\textwidth]{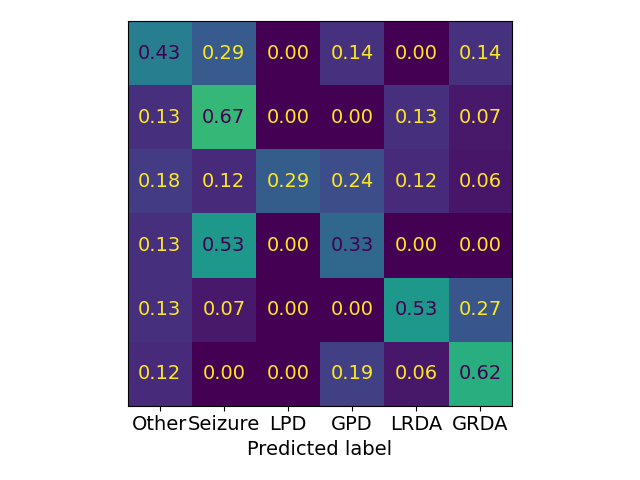}
         \caption{User 6 without AI.}
    \end{subfigure}
    \hspace{-3.35em}
    \begin{subfigure}[b]{0.29\textwidth}
         \centering
         \includegraphics[width=\textwidth]{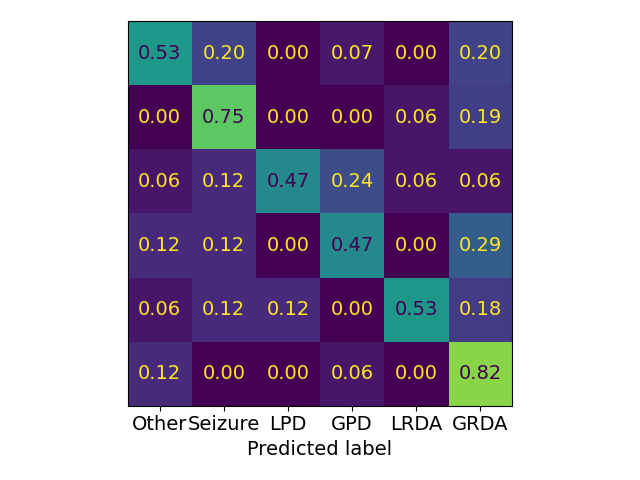}
         \caption{User 6 with AI.}
    \end{subfigure}
    \newline
    \begin{subfigure}[b]{0.29\textwidth}
         \includegraphics[width=\linewidth]{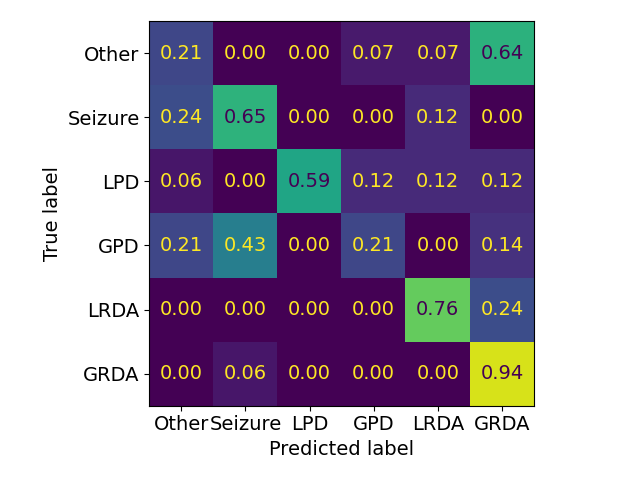}
        \caption{User 7 without AI.}
    \end{subfigure}
    \hspace{-2.8em}
    \begin{subfigure}[b]{0.29\textwidth}
         \centering
         \includegraphics[width=\textwidth]{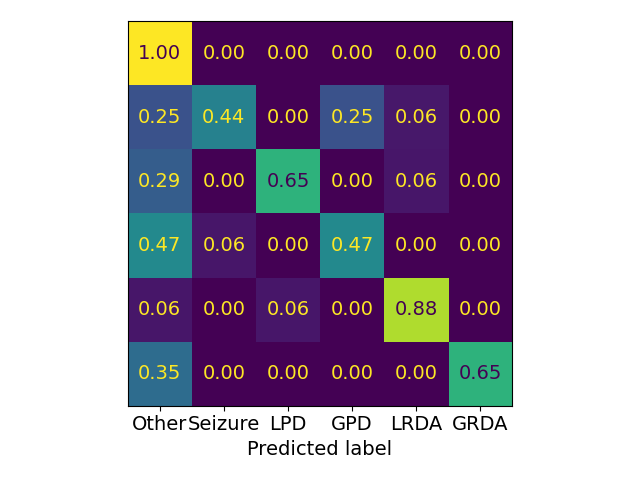}
         \caption{User 7 with AI.}
    \end{subfigure}
    \hspace{-3.35em}
    \begin{subfigure}[b]{0.29\textwidth}
         \centering
         \includegraphics[width=\textwidth]{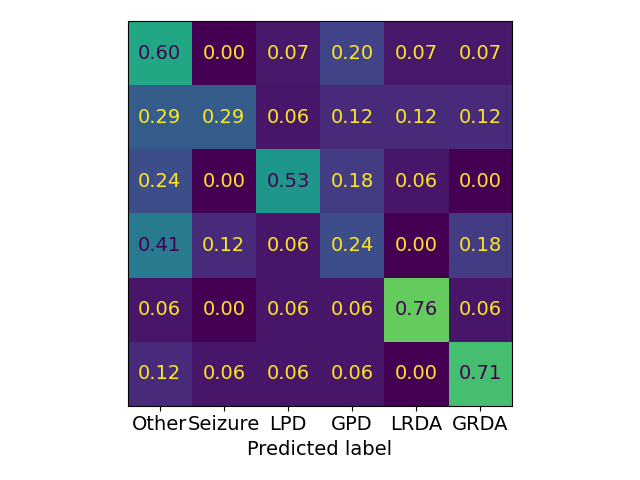}
         \caption{User 8 without AI.}
    \end{subfigure}
    \hspace{-3.35em}
    \begin{subfigure}[b]{0.29\textwidth}
         \centering
         \includegraphics[width=\textwidth]{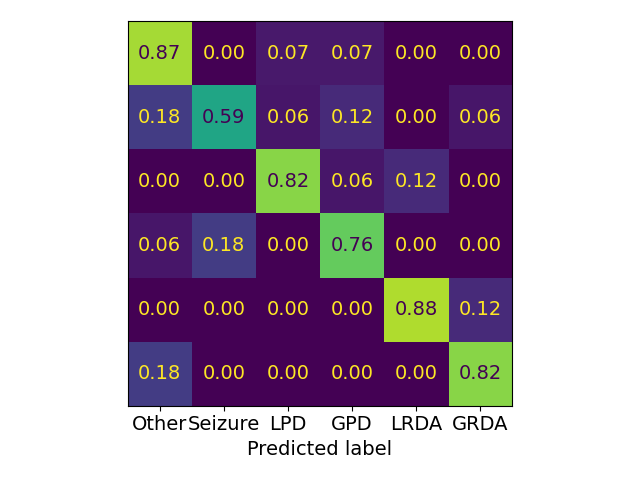}
         \caption{User 8 with AI.}
    \end{subfigure}
  \caption{The error matrices for each user with and without AI.}
  \label{fig:conf_matrices}
\end{figure}

\FloatBarrier
In Figure \ref{fig:conf_matrices}, we show the error matrices of each user with and without AI.

\FloatBarrier
\section{Extended Analysis for External Validation}\label{app:ext_val}
\FloatBarrier
To account for the large class distribution shift between the internal (original, n=35,740) and external (new, n=1,500) test datasets, we use bootstrapping to sample the external dataset such that the class distributions match.

We report the mean and standard deviation for 1000 bootstrap samples. We use majority-vote (for each sample, the class for which the most annotators voted) as the ground truth for this analysis. To construct a bootstrap sample for the external dataset, we sample with replacement for each class, matching the class distribution of the internal dataset. To construct a bootstrap sample for the internal dataset, we sample with replacement matching the class distribution of the internal dataset.


\FloatBarrier
\section{Statistical Significance and Uncertainty Calculations}\label{app:sig}
\FloatBarrier
\subsection{Uncertainty Calculation for AUPRC and AUROC using bootstrapping}\label{app:boot}
We take $N=1000$ bootstrap samples of size $|D^t|$ samples with replacement from the test set, where $|D^t|$ is the size of the test set. We calculate the AUPRC and AUROC for each bootstrap sample, presenting the median AUPRC or AUROC in Table \ref{tab:aucs}. 
The 95\% CI is calculated by taking metrics of the 2.5 percentile and 97.5 percentile of the bootstrap samples. 
\FloatBarrier

\subsection{Bootstrap by Patient}

We conducted an experiment with an additional bootstrap method where each bootstrap sample consists of the EEG samples from 761 randomly selected patients with replacement from the test set (the test set has 761 patients). Refer to Table \ref{tab:aucs_patient_bootstrap} for values. Whether bootstrapping by patient, as shown here, or by sample, as shown in the main text, our interpretable method is statistically significantly better than the SPaRCNet \citet{older_eeg_paper}.

\begin{table*}[ht]
  \caption{AUROC and AUPRC of the interpretable model compared to SPaRCNet from \citet{older_eeg_paper}. Here we present the results of bootstrapping by patient.}
  \label{tab:aucs_patient_bootstrap}
  \scriptsize
  \begin{tabular}{llccccccc}
    \toprule
    & & Other & Seizure & LPD & GPD & LRDA & GRDA & All\\
    \midrule
    \multirow{2}{1cm}{AUROC} & Interp. & \textbf{0.80 [0.76, 0.84]}& \textbf{0.87 [0.83, 0.91]}& \textbf{0.93 [0.90, 0.95]}& \textbf{0.95 [0.92, 0.98]}& \textbf{0.92 [0.88, 0.96]}& \textbf{0.93 [0.90, 0.95]}& \textbf{0.90 [0.88, 0.93]}\\
    & Uninterp. \cite{older_eeg_paper} & 0.79 [0.79, 0.80]& 0.86 [0.85, 0.88]& 0.90 [0.90, 0.90] & 0.94 [0.94, 0.95]& 0.92 [0.92, 0.92]  & 0.92 [0.91, 0.92]& 0.89 [0.89, 0.90]\\
    \multirow{2}{1cm}{AUPRC} & Interp. & \textbf{0.53 [0.47, 0.58]}& \textbf{0.26 [0.16, 0.37]}& \textbf{0.81 [0.73, 0.86]}& \textbf{0.91 [0.82, 0.96]}& \textbf{0.75 [0.59, 0.85]}& \textbf{0.67 [0.57, 0.75]}& \textbf{0.74 [0.68, 0.79]}\\
    & Uninterp. \cite{older_eeg_paper} & 0.47 [0.46, 0.48]& 0.19 [0.16, 0.23]& 0.73 [0.73, 0.74]& 0.89 [0.88, 0.89]&  0.74 [0.73, 0.74]& 0.63 [0.61, 0.64]& 0.70 [0.69, 0.70]\\
    \bottomrule
  \end{tabular}
\end{table*}

To further test the significance in difference between our interpretable model and the uninterpretable model's bootstrapped results, we also calculated the percentage of bootstrap samples where interpretable model performs better than the uninterpretable model of \citet{older_eeg_paper}, shown in Table \ref{tab:significance_test_percent}.
\begin{table*}[ht]
  \centering
  \caption{The percentage of bootstrap samples where interpretable model performs better than the uninterpretable model of \citet{older_eeg_paper}.}
  \label{tab:significance_test_percent}
  \scriptsize
  \begin{tabular}{llccccccc}
    \toprule
    & & Other & Seizure & LPD & GPD & LRDA & GRDA & All\\
    \midrule
    \multirow{2}{1cm}{AUROC} & Sample Bootstrap & 98.90 & 80.60 & 100.00 & 100.00 & 97.90 & 100.00 & 100.00   \\
    & Patient Bootstrap & 62.4  & 70.40  & 96.40 & 77.40 & 60.00 & 88.10  & 77.00 \\
    \multirow{2}{1cm}{AUPRC} & Sample Bootstrap & 100.00 & 98.00 & 100.00 & 100.00 & 100.00 & 100.00 &  100.00 \\
    & Patient Bootstrap & 98.30 & 86.70  & 97.20 & 68.50 & 64.00 & 82.00  & 93.60 \\
    \bottomrule
  \end{tabular}
\end{table*}

\FloatBarrier
\subsection{p-values for Neighborhood Analyses}
Additional exploration of the statistical significance of our neighborhood analysis results can be found in Table \ref{tab:significance_test}. The interpretable model is always significantly better than the uninterpretable model with p-values much lower than 0.001.

\begin{table*}[ht]
  \centering
  \caption{Here we present the significance (p-values) of the neighborhood analysis results of our interpretable model vs. the uninterpretable model of \citet{older_eeg_paper}. The interpretable model was always significantly better.}
  \label{tab:significance_test}
  \scriptsize
  \begin{tabular}{llccccccc}
    \toprule
    & Other & Seizure & LPD & GPD & LRDA & GRDA & All\\
    \midrule
    Neighborhood Analysis by Max & 3.13e-45 & 2.26e-12 & 6.68e-128 & 1.23e-68 & 3.64e-134 & 4.13e-32 & 0.00   \\
    Neighborhood analysis by Vote & 3.75e-124 & 3.47e-09 & 1.10e0159 & 2.78e-84 & 6.41e-132 & 4.29e-71 &  0.00 \\
    \bottomrule
  \end{tabular}
\end{table*}

\subsection{Reproducibility of the {\tt Annoy} package for neighborhood analysis calculation}

Due to the large amount of data in our study, traditional nearest-neighbor calculations are intractable and we must use approximation code. The state-of-the-art Annoy package \cite{annoy_spotify} is used to find approximate nearest neighbors for our neighborhood analysis metrics. We used the high-computation cost, highest-accuracy setting on this package to extract the nearest neighbors for each sample for neighborhood analysis. In its current implementation, the Annoy approximation algorithm is not fully reproducible, which is a finding brought up by multiple other users. To account for this issue, we ran the algorithm multiple times and compared results, finding that neighborhood analysis results varied by less than 0.01\%.

\FloatBarrier
\section{Neighborhood Analyses}\label{app:neighbor_analyses}
\FloatBarrier

\begin{figure}
  \centering
  \includegraphics[width=\linewidth]{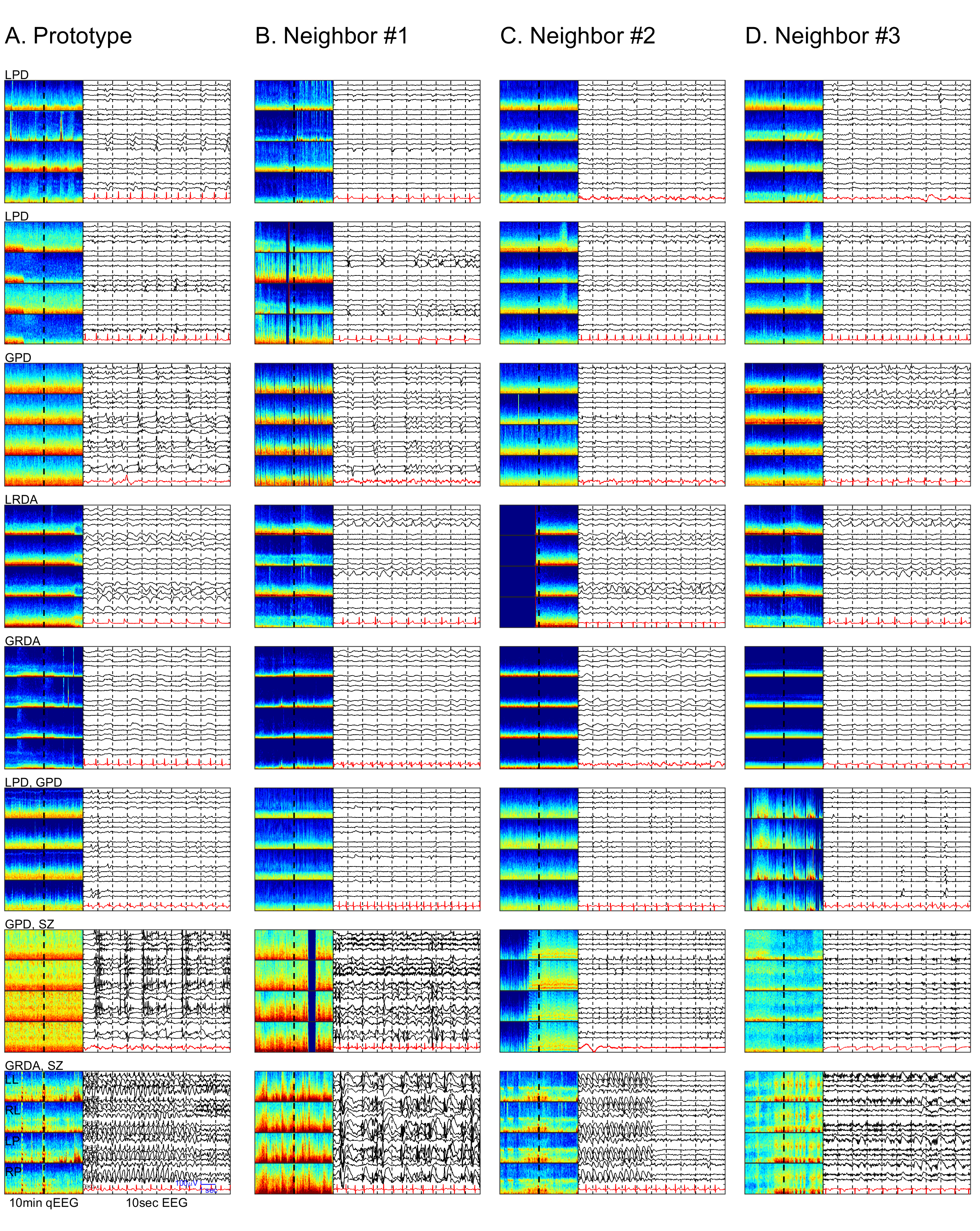}
  \caption{The nearest neighbors for a prototype of the ProtoPMed-EEG model. We show top 3 nearest test samples corresponding to the prototype in 8 cases. The full set of prototypes and their neighbors can be found at the link in Appendix \ref{app:neighbor_analyses}.}
  \label{fig:neighborhood}
\end{figure}

In Figure \ref{fig:neighborhood}, for each of the eight prototypes from our model, we explore the three nearest neighbors to the prototype from the test set. In each case, the neighboring samples are similar to the prototype not only in class but also in amplitude, peak-to-peak distance and other domain-relevant qualities. This demonstrates to domain experts our model's concept of ``similarity.'' Qualitative neighborhood analyses for all prototypes showing the six nearest neighbors from each of the training and test sets can be found at \url{https://drive.google.com/file/d/1AFEt5IqPiSWLafHE6j1tl1_XSBTj2h0f/view?usp=share_link}.

\FloatBarrier
\section{Dataset} \label{app:data}
    The method for gathering and labelling data is described in \citet{iir_eeg_paper} and \citet{older_eeg_paper}. We use Training Dataset 3 from  \citet{iir_eeg_paper} as our training dataset for ProtoPMed-EEG. The model weights used to initialize our model, from Jing et al. \citet{older_eeg_paper}, are trained on Training Datasets 1, 2, and 3. The test set for both models is Test Dataset 4.

\section{Videos link}\label{app:vid_links}
\url{https://warpwire.duke.edu/w/8zoHAA/}

\section{Code link}\label{app:code_links}
\url{https://github.com/chengstark/Interp.-EEG-Public}

\FloatBarrier
\FloatBarrier

\end{document}